\documentclass[10pt,twocolumn,letterpaper]{article}

\usepackage[pagenumbers]{cvpr}      

\usepackage{multirow}
\usepackage{placeins}
\usepackage{amsmath,amsfonts}
\usepackage{algorithmic}
\usepackage{array}
\usepackage{xcolor}
\usepackage{wrapfig}

\definecolor{pastelblue}{RGB}{100,170,210}  
\definecolor{pastelred}{RGB}{240,120,150}   

\definecolor{pastelgreen}{RGB}{120,200,120}  
\definecolor{pastelyellow}{RGB}{255,230,120}  
\definecolor{pastelpurple}{RGB}{180,130,210}  
\definecolor{pastelorange}{RGB}{255,160,100}  
\usepackage{textcomp}
\usepackage{stfloats}
\usepackage{url}
\usepackage{verbatim}
\usepackage{graphicx}
\def\BibTeX{{\rm B\kern-.05em{\sc i\kern-.025em b}\kern-.08em
    T\kern-.1667em\lower.7ex\hbox{E}\kern-.125emX}}
\usepackage{balance}
\usepackage{pgfplots}
\pgfplotsset{compat=1.18} 
\usepackage{filecontents} 
\usepgfplotslibrary{fillbetween}

\usepackage{float}
\usepackage{booktabs}
\usepackage{adjustbox}

\usepackage{colortbl}
\usepackage{xcolor}
\usepackage{pifont}
\usepackage{booktabs}
\usepackage{array}
\usepackage{graphicx}
\usepackage{makecell}

\definecolor{forestgreen}{RGB}{34,139,34}
\newcommand{\cmark}{\textcolor{forestgreen}{\ding{51}}} 
\newcommand{\xmark}{\textcolor{red}{\ding{55}}} 

\definecolor{cvprblue}{rgb}{0.21,0.49,0.74}
\usepackage[pagebackref,breaklinks,colorlinks,allcolors=cvprblue]{hyperref}

\def\paperID{17797} 
\def\confName{CVPR}
\def\confYear{2026}

\title{KeyPointDiffuser: Unsupervised 3D Keypoint Learning via Latent Diffusion Models}

\author{
Rhys Newbury${^{1,2}}$, Juyan Zhang${^1}$, Tin Tran${^1}$,  Hanna Kurniawati${^2}$, Dana Kuli\'{c}${^1}$\\
${^1}$Monash University, ${^2}$ Australian National University`\\
{\tt\small \{rhys.newbury, juzyan.zhang, trung.tran, dana.kulic\}@monash.edu}
\\{\tt\small \{hanna.kurniawati\}@anu.edu.au}
}

\begin{document}
\maketitle

\begin{abstract}
Understanding and representing the structure of 3D objects in an unsupervised manner remains a core challenge in computer vision and graphics. Most existing unsupervised keypoint methods are not designed for unconditional generative settings, restricting their use in modern 3D generative pipelines; our formulation explicitly bridges this gap. We present an unsupervised framework for learning spatially structured 3D keypoints from point cloud data. These keypoints serve as a compact and interpretable representation that conditions an Elucidated Diffusion Model (EDM) to reconstruct the full shape. The learned keypoints exhibit repeatable spatial structure across object instances and support smooth interpolation in keypoint space, indicating that they capture geometric variation. Our method achieves strong performance across diverse object categories, yielding a 6 percentage-point improvement in keypoint consistency compared to prior approaches.
\end{abstract}

\section{Introduction}

Understanding the structural semantics of 3D objects is a fundamental challenge in computer vision and graphics. A key step toward this goal is the discovery of consistent and interpretable \textit{keypoints}, i.e., sparse sets of points that capture salient object features and enable tasks such as correspondence, reconstruction, and manipulation. While a wide range of methods have been developed for 2D keypoint discovery, including both handcrafted~\cite{harris1988combined} and learning-based approaches~\cite{lowe2004distinctive}, extending these ideas to the 3D domain remains significantly less explored~\cite{shi2021skeletonmerger, zhu20233dkeypointestimationusing, jakab2021keypointdeformer, NEURIPS2024_582e9771}.

Unsupervised keypoint discovery aims to learn meaningful object structures without relying on labeled data, simplifying the training process and improving scalability~\cite{you2022keypointdiscovery, zhu20233dkeypointestimationusing, NEURIPS2024_582e9771, shi2021skeletonmerger}. In this paper, we introduce \textbf{KeyPointDiffuser}, a novel \textit{unsupervised} framework for learning 3D keypoints by leveraging latent diffusion models. Our key insight is to treat the extracted keypoints as a latent representation that conditions the generation of full 3D shapes through a denoising diffusion process. This generative formulation enables the model to learn structural priors from data, allowing it to extract semantically consistent and spatially informative keypoints \textit{without supervision}.

\begin{figure*}[t]
    \centering
    \includegraphics[width=0.8\linewidth]{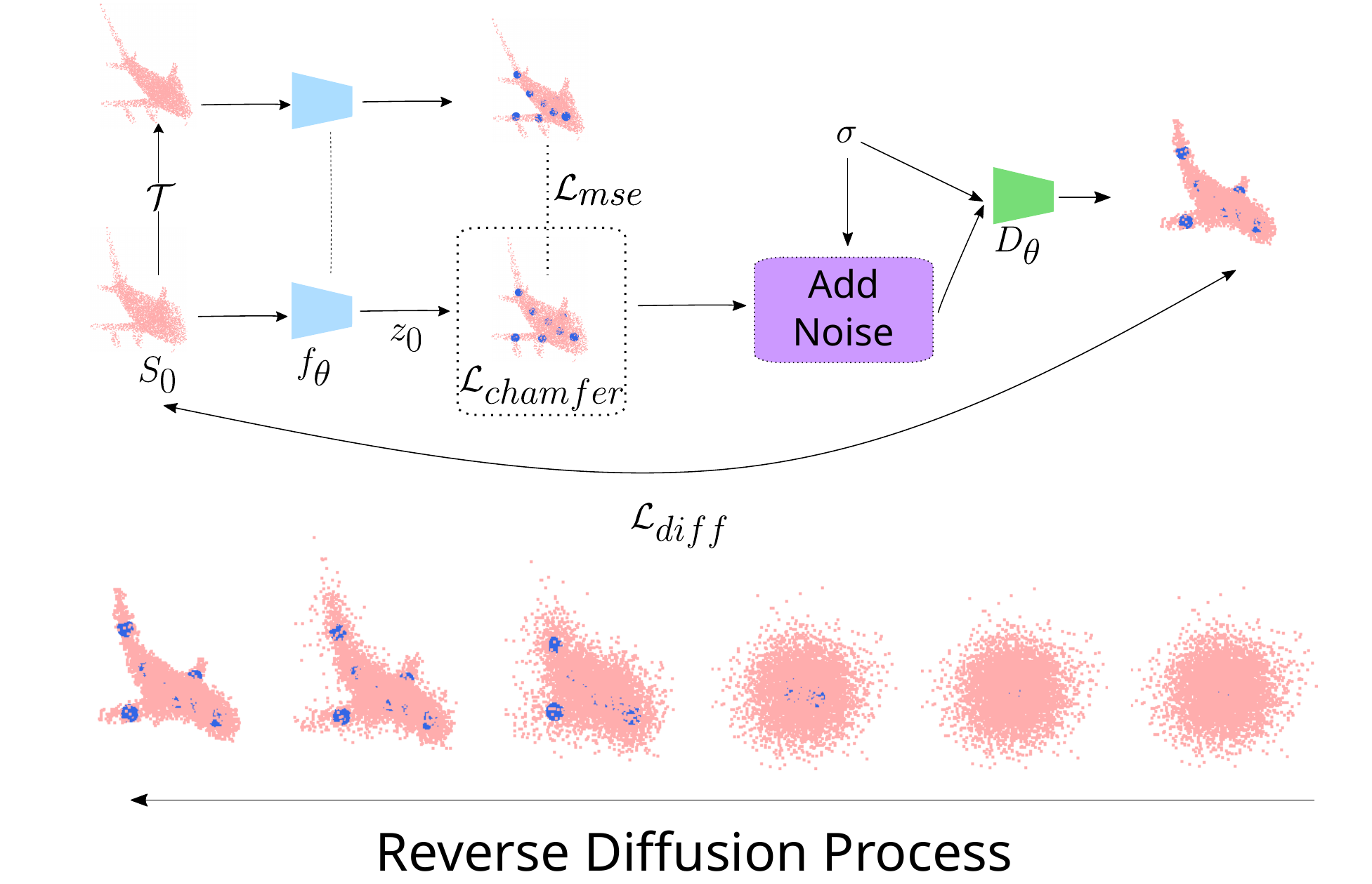}
    \caption{\textbf{Top: } Overview of the keypoint-conditioned 3D shape generation pipeline. The input point cloud $S_0$ is encoded into a structured latent code $z_0 = z_{\text{kp}} \oplus z_{\text{aux}}$, where $z_{\text{kp}}$ denotes the set of learned 3D keypoints and $z_{\text{aux}}$ represents auxiliary latent features sampled from a Gaussian distribution. These keypoints guide a denoising diffusion model to iteratively reconstruct the original shape. The keypoints are regularized using a Chamfer loss $\mathcal{L}_{\text{chamfer}}$ and a deformation consistency loss $\mathcal{L}_{\text{mse}}$, where $\mathcal{T}$ denotes a differentiable geometric transformation applied to the input shape to simulate structured deformations (e.g., stretching, bending, twisting, tapering).
 \textbf{Bottom:} The reverse diffusion process refines the noisy input to produce a plausible shape consistent with the keypoints. Since the noise is sampled from a standard range ($[-1, 1]$), but diffused shapes can occupy smaller spatial extents, the process appears to "zoom in" as noise is removed, causing keypoints (blue circles) to emerge and grow more prominent over time. Timesteps are sampled on a logarithmic scale.}
    \label{fig:overview}
\end{figure*}

In parallel, diffusion-based generative models have recently achieved impressive results in 3D shape generation~\cite{luo2021diffusionprobabilisticmodels3d, zeng2022liondiffusion}, which inspires our approach. Our work aims to bridge these domains by introducing a latent diffusion model \textit{conditioned on learned keypoints} for unsupervised 3D keypoint discovery. Unlike prior works that rely on explicit structural priors or deterministic decoders~\cite{shi2021skeletonmerger, zhu20233dkeypointestimationusing, jakab2021keypointdeformer}, we leverage a probabilistic generative decoder that learns to reconstruct shapes from keypoints directly. An overview of our keypoint-conditioned generation pipeline is illustrated in Figure~\ref{fig:overview}.

We demonstrate that KeyPointDiffuser produces keypoints that are both geometrically meaningful and semantically aligned across diverse object categories, as evidenced by consistently higher keypoint correlation scores. Notably, our method outperforms state-of-the-art methods~\cite{NEURIPS2024_582e9771,jakab2021keypointdeformer,SC3K,shi2021skeletonmerger} on 8 out of 13 object categories for the dual alignment score (DAS) metric, and 8 out of 9 categories for keypoint correlation metric. In addition, the compact and informative nature of our keypoint-based latent representation allows us to generate new objects from a set of keypoints. Empirically our diffusion model generates shapes whose distribution more closely aligns with the ground truth data compared to both KeypointDeformer~\cite{jakab2021keypointdeformer} and the point cloud diffusion model (DPM)~\cite{luo2021diffusionprobabilisticmodels3d} across 7 of 13 classes. These results validate that the learned keypoints are not only interpretable but also effective for guiding shape reconstruction and supporting downstream applications.

Our main contributions are as follows:
\begin{itemize}
    \item We propose a generative formulation that leverages denoising diffusion to learn compact and interpretable 3D keypoints without any ground-truth supervision.
    \item We introduce a structured latent space composed of explicit keypoints and auxiliary features, and enforce geometric regularization techniques, that shapes the latent space to align with the ELBO objective, promoting spatially meaningful, consistent keypoints and improved generative fidelity.   
    \item We demonstrate that our approach achieves strong keypoint consistency and enables high-fidelity shape generation.
    \item We provide a unified training framework for testing keypoint discovery algorithms.
\end{itemize}

\section{Related Work}

\subsection{Unsupervised 2D Keypoint Discovery}

Unsupervised methods for 2D keypoint discovery aim to learn compact and interpretable object representations without supervision. Early approaches enforced equivariance, ensuring keypoints transform consistently under known perturbations \cite{Thewlis2017}. Others used reconstruction bottlenecks, where images are generated from a small set of keypoints, encouraging the emergence of meaningful landmarks \cite{Zhang2018, Locatello2020}. A related direction disentangles shape from appearance using part-based representations \cite{sandro2021, Shu2018}. These methods increasingly produce robust, semantically consistent keypoints that align with human-interpretable object parts, demonstrating the viability of keypoints as structured, low-dimensional representations.

\subsection{Low-Dimensional Representations for 3D Reconstruction}

3D reconstruction methods often compress geometry into latent vectors \cite{Wu2016, Achlioptas2018}, though such embeddings lack interpretability. Structured alternatives like keypoints \cite{Suwajanakorn2018, Novotny2019} or skeletons \cite{SkeletonNet} provide more intuitive control, enabling tasks like pose estimation and shape manipulation. Parametric templates, such as 3D morphable face models \cite{Blanz1999} and SMPL for human bodies \cite{Loper2015}, define low-dimensional spaces for shape variation, though often category-specific. Other works combine deep learning with structured representations \cite{Sahasrabudhe2019}, improving both control and fidelity. 

\subsection{3D Keypoint Learning}

\begin{table}[h]
\centering
\begin{adjustbox}{max width=\linewidth}
\begin{tabular}{l c c c c}
\toprule
\textbf{Method} & 
\makecell{\textbf{Keypoint}\\\textbf{Discovery}} &
\textbf{Reconstruction} &
\makecell{\textbf{Reference}\\\textbf{Based Generation}} &
\makecell{\textbf{Reference}\\\textbf{Free Generation}} \\
\midrule
KeypointDeformer~\cite{jakab2021keypointdeformer} & \cmark & \cmark & \cmark & \xmark \\
Key-Grid~\cite{NEURIPS2024_582e9771}           & \cmark & \cmark & \xmark & \xmark \\
Skeleton Merger~\cite{shi2021skeletonmerger} & \cmark & \cmark & \xmark & \xmark \\
SC3K~\cite{SC3K} & \cmark & \xmark & \xmark & \xmark \\
\textbf{Ours}                         & \cmark & \cmark & \xmark & \cmark \\
\bottomrule
\end{tabular}
\end{adjustbox}
\caption{Comparison of methods across key capabilities. Our approach can generate objects using the sampled keypoints without extra information such as a reference mesh.}
\label{tab:approaches}
\end{table}

Unsupervised 3D keypoint learning seeks to discover consistent keypoints either with or without annotations. Early works used autoencoders to reconstruct shapes from sparse keypoints, often guided by geometric priors such as symmetry \cite{fernandez2020unsupervised}, skeletal structures \cite{shi2021skeletonmerger}, or cage-based deformation \cite{jakab2021keypointdeformer}. More recent methods explore implicit or adversarial objectives to improve robustness and generalization \cite{you2022keypointdiscovery, zhu20233dkeypointestimationusing} by enforcing various invariances for consistency. SC3K~\cite{SC3K} ensures keypoint stability under rotation/noise, while SelfGeo~\cite{selfgeo} enforces geodesic consistency for deformable shapes. However, most unsupervised approaches still rely on handcrafted priors or deterministic decoders. By contrast, supervised frameworks like KeypointDETR \cite{KeypointDETR} achieve strong performance with labeled keypoints (at the expense of annotation effort). Recent unsupervised approaches address these limitations by leveraging more expressive models. For example, Key-Grid \cite{NEURIPS2024_582e9771} uses a grid-based heatmap decoder to improve semantic consistency across both rigid and deformable shapes, while our method employs a latent diffusion model conditioned on learned keypoints for flexible, probabilistic shape reconstruction without explicit structural constraints. A summary of the unsupervised approaches is provided in Table \ref{tab:approaches}

\section{Methodology}

We propose an unsupervised approach for learning keypoints from 3D shapes. Our method leverages a latent diffusion model to extract semantically consistent keypoints. 


\subsection{Unsupervised Keypoint Learning}
\label{sec:methodology}

Given a 3D shape represented as a point cloud $S_0 \in \mathbb{R}^{N \times 3}$, where $N$ is the number of points in the shape, our model aims to learn a structured latent representation $z_0 \in \mathbb{R}^{d \times 3 + m}$, which consists of 3D keypoints represented as $K \in \mathbb{R}^{d \times 3}$ for $d$ keypoints and an auxiliary feature vector $z_{\text{aux}} \in \mathbb{R}^{m}$.

\paragraph{Keypoint Extraction and Shape Encoding}  

\begin{figure}
    \centering
    \includegraphics[width=0.75\linewidth, trim={5cm 5cm 5cm 5cm}, clip]{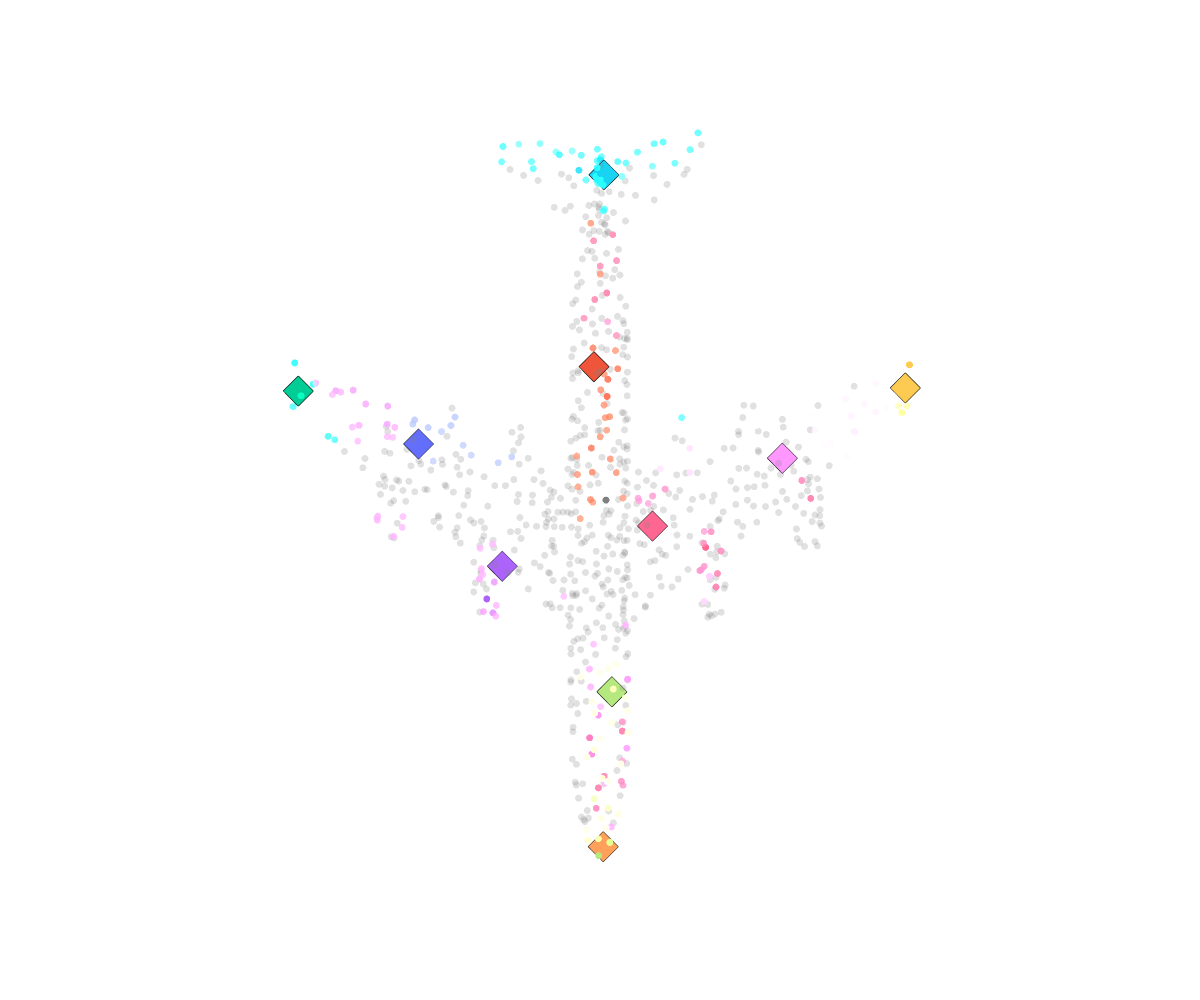}
    \caption{Visualization of the model’s keypoints (colored diamonds) and their corresponding high-attention regions (colored points) overlaid on the input point cloud (gray). Each color represents a distinct keypoint and its associated attention distribution.}
    \label{fig:dist}
\end{figure}

The input point cloud $S_0$ is encoded by a transformer-based encoder $F_{\theta}$, which extracts a set of geometric keypoints and an auxiliary latent vector. Specifically, the encoder first computes per-point features using a point cloud transformer backbone\cite{wu2024pointtransformerv3simpler}. Following \citet{shi2021skeletonmerger, NEURIPS2024_582e9771}, we learn a weighting matrix over input points. However, their method predicts weights through a dedicated weighting module, we obtain them directly via multi-head attention applied to the transformer features. Each learnable keypoint query attends to the point set, yielding a per-point attention distribution which are normalized using softmax across all input points. An example of the learned keypoint attention distribution is shown in Fig \ref{fig:dist}. We then form each keypoint as an attention-weighted convex combination of the input coordinates, producing $K \in \mathbb{R}^{d \times 3}$. This ensures all keypoints are inside the convex hull of the point cloud. Formally, for each keypoint $k \in \{1, \dots, d\}$, the encoder predicts a set of attention weights
$a_k = (a_{k1}, a_{k2}, \dots, a_{kN})$ satisfying
\begin{equation}
a_{ki   } \ge 0, \quad \sum_{i=1}^N a_{ki} = 1.
\end{equation}
The keypoint is then obtained as a convex combination of the input coordinates:
\begin{equation}
K_k = \sum_{i=1}^N a_{ki} \, x_i.
\end{equation}
Since each $a_k$ lies on the probability simplex ($a_{ki} \ge 0, \sum_i a_{ki} = 1$),
each keypoint $K_k$ is guaranteed to lie within the convex hull of the input point cloud:
\begin{equation}
K_k \in \mathrm{conv}\{ x_1, x_2, \dots, x_N \}, \quad \forall k \in \{1, \dots, d\}.
\end{equation}


Simultaneously, the encoder aggregates the attention outputs across all keypoints to produce a global descriptor $h_{\text{aux}}$, which captures unstructured, shape-specific variations. The auxiliary features $z_{\text{aux}}$ are modeled as a Gaussian distribution parameterized by a mean and log-variance predicted by a small neural network $\phi$ given $h_{\text{aux}}$. The two components $(K, z_{\text{aux}})$ together form the structured and unstructured parts of the latent representation. During training, sampling is performed using the reparameterization trick to allow backpropagation:

\begin{align}
    (K, h_{\text{aux}}) &= F_{\theta}(S_0), 
        && K \in \mathbb{R}^{d \times 3} \label{eq:encoder-output-new} \\
    (\mu, \log \sigma^2) &= \phi(h_{\text{aux}}),
        && \mu, \log \sigma^2 \in \mathbb{R}^{m} \label{eq:gaussian-params-new} \\
    z_{\text{aux}} &= \mu + \sigma \odot \epsilon,
        && \epsilon \sim \mathcal{N}(0, I), \quad z_{\text{aux}} \in \mathbb{R}^{m} \label{eq:reparam-new}
\end{align}

To ensure the predicted keypoints lie on the surface of the input shape, we introduce a differentiable \emph{soft projection} mechanism that aligns each keypoint with the object point cloud. This aims to guarantee that surface geometry information is consistently accessible to the decoder throughout training, including when keypoint predictions are unreliable. This projection is not used in any of the keypoint related loss terms, it is used only to condition the decoder with surface-aligned keypoints. A full definition of this mechanism is provided in Appendix~\ref{app:softproject}.


Finally, the keypoints and stochastic auxiliary code are concatenated to form the complete latent vector:
\begin{align}
    z_0 &= \mathrm{vec}(K) \oplus z_{\text{aux}},
        && z_0 \in \mathbb{R}^{3d + m}, \label{eq:latent-concat-new}
\end{align}
where $\mathrm{vec}(\cdot)$ denotes vectorization and $\oplus$ denotes concatenation. This composite latent code $z_0$ provides both \emph{explicit geometric structure} through $K$ and \emph{implicit variation} through $z_{\text{aux}}$, and is used to condition the generative diffusion model for 3D shape reconstruction.

\paragraph{Shape Diffusion Process}

To generate a dense shape from the extracted keypoints, we employ the methodology presented by Elucidated Diffusion Model (EDM)~\cite{karras2022elucidating} described in Appendix~\ref{app:edm_forward}.

However, to improve stability and efficiency during training, we employ a \textit{curriculum-based noise scheduling strategy}. Specifically, we progressively adjust the log-normal distribution from which the noise scale $\sigma$ is sampled. Early in training, smaller $\sigma$ values are emphasized to allow the model to learn coarse structure; as training progresses, the noise distribution shifts toward higher variance, encouraging robustness to increasingly noisy inputs.

This is implemented via an evolving Gaussian distribution over $\log \sigma$, where both the mean and standard deviation are linearly interpolated over the initial 80\% of training time:

\begin{equation}
    \ln(\sigma) \sim \mathcal{N}(\mu_n(e), \sigma_n^2(e))
\end{equation}

with

\begin{align}
    \mu_n(e) &= (1 - \alpha(e)) \cdot \mu_{\text{init}} + \alpha(e) \cdot \mu_{\text{final}} \\
    \sigma_n(e) &= (1 - \alpha(e)) \cdot \sigma_{\text{init}} + \alpha(e) \cdot \sigma_{\text{final}}
\end{align}

and $\alpha(e) = \text{min}(\frac{e}{0.8E},1)$ denoting the normalized training progress, i.e. current iteration $e$ divided by the total number of iterations $E$. The curriculum is applied only during the first 80\% of training, ensuring that the model is exposed to the full task difficulty during the final 20\%.


\paragraph{Shape Reconstruction Loss.}

Our reconstruction objective is based on the reverse diffusion dynamics of the Elucidated Diffusion Model (EDM) framework~\cite{karras2022elucidating}, adapted to operate on point clouds. Given a clean point cloud $S_0 \in \mathbb{R}^{N \times 3}$ and a noise level $\sigma_t$, we construct a noisy input
\begin{equation}
S_t = S_0 + \sigma_t \epsilon, \quad \epsilon \sim \mathcal{N}(0, I),
\end{equation}
and train a denoising network $D_\theta$ to predict a cleaned reconstruction $\hat{S} = D_\theta(S_t, \sigma_t, z_0)$ conditioned on both the noise level $\sigma_t$ and the latent keypoint code $z_0$.
Following EDM, we supervise the model across a range of noise levels using a noise-dependent weight $w(\sigma_t)$ that balances gradient contributions across $\sigma_t$.

Rather than using a purely symmetric Chamfer Distance, we employ an \emph{asymmetric} variant that separates precision and coverage. 
Given two point clouds $A = \{a_i\}_{i=1}^M$ and $B = \{b_j\}_{j=1}^N$, we define the one-way Chamfer distance
\begin{equation}
d_{\text{CD}}(A \rightarrow B) 
= \frac{1}{|A|} \sum_{a \in A} \min_{b \in B} \| a - b \|_2^2,
\label{eqn:chamfer_distance_general}
\end{equation}
which measures how well points in $A$ are explained by points in $B$.
We then form an asymmetric reconstruction loss
\begin{equation}
\mathcal{L}_{\text{CD}}(\hat{S}, S_0)
=
\alpha \, d_{\text{CD}}(\hat{S} \rightarrow S_0)
\;+\;
\beta \, d_{\text{CD}}(S_0 \rightarrow \hat{S}),
\label{eq:asym_chamfer}
\end{equation}
where the first term encourages \emph{precision} (predicted points lie on the target surface) and the second term encourages \emph{coverage} (all regions of the target surface are represented by at least one predicted point). 
We choose $\beta > \alpha$ to penalize missing regions more strongly than over-concentrated regions.

To further discourage point collapse and promote uniform spatial coverage, we introduce a local repulsion term that penalizes clusters of predicted points that are too close to each other.
Let $\mathcal{N}_k(i)$ denote the $k$ nearest neighbors of a predicted point $\hat{x}_i \in \hat{S}$.
We define
\begin{equation}
\mathcal{L}_{\text{repel}}(\hat{S}) 
=
\frac{1}{N k}
\sum_{i=1}^{N}
\sum_{j \in \mathcal{N}_k(i)}
\max \Big( 0,\; m - \| \hat{x}_i - \hat{x}_j \|_2 \Big),
\label{eq:repulsion}
\end{equation}
where $m$ is a margin that sets the minimum desired spacing. 
This hinge penalty activates only when two predicted points lie within distance $m$, encouraging them to spread out and avoiding overly dense point clusters.

Finally, we combine these terms into a single diffusion training objective:
\begin{align}
\mathcal{L}_{\text{diff}}
&=
\mathbb{E}_{\sigma_t, S_0, \epsilon}
\left[
w(\sigma_t)
\cdot
\mathcal{L}_{\text{CD}}\big(D_\theta(S_0 + \sigma_t \epsilon, \sigma_t, z_0),\, S_0\big)
\right]
\\ &+ \rho \,
\mathbb{E}_{\sigma_t}
\left[
\gamma(\sigma_t) \,
\mathcal{L}_{\text{repel}}\big(D_\theta(S_0 + \sigma_t \epsilon, \sigma_t, z_0)\big)
\right], \nonumber
\label{eq:full_loss}
\end{align}
where $\rho$ controls the strength of the repulsion term and 
$\gamma(\sigma_t) = \frac{\sigma_{\text{data}}}{\sigma_t + \sigma_{\text{data}}}$ 
down-weights the repulsion penalty at high noise levels and up-weights it at low noise levels.
Intuitively, the first expectation trains the model to denoise across a range of noise scales in the EDM sense, while the second expectation explicitly encourages the final reconstructions (low $\sigma_t$) to be both complete and evenly distributed over the surface, rather than collapsing into a few high-density regions.

Appendix~\ref{app:shape_edm} provides implementation details for $w(\sigma_t)$, the sampling schedule over $\sigma_t$, and the diffusion-based shape decoding process.

\paragraph{Chamfer Loss for Keypoint Alignment}  

To encourage the predicted keypoints to lie close to the underlying input geometry, we define a loss based on the one-way Chamfer Distance between the keypoints $K \in \mathbb{R}^{d \times 3}$ and the input point cloud $S_0 \in \mathbb{R}^{N \times 3}$, $\mathcal{L}_{\text{chamfer}} = d_{\text{CD}}(K \rightarrow S_0)$. This formulation penalizes keypoints that deviate from the input surface, while the furthest point sampling-based loss should encourage coverage of the object.

\paragraph{Deformation Consistency for Keypoint Robustness}  

To encourage keypoints that are semantically meaningful and consistent across object instances, we introduce a deformation consistency loss. Given an input shape $S_0$, we apply a structured deformation $\mathcal{T}$ to obtain a transformed shape, $S_d = \mathcal{T}(S_0)$, where $\mathcal{T}$ is a differentiable and semantically meaningful transformation that preserves structural validity. While it can encompass rigid, non-rigid, or articulated deformations depending on context, we focus on a subset that simulates plausible intra-class variations, specifically stretching, bending, twisting, and tapering  (see Appendix~\ref{appendix:deformations}).

Keypoints are extracted from both the original and deformed shapes:

\begin{equation}
K = F_{\theta}(S_0), \quad K_d = F_{\theta}(S_d),
\end{equation}

and consistency is enforced by minimizing the mean squared error between transformed original keypoints and those from the deformed shape. We use MSE instead of Chamfer Distance, as keypoints are ordered and should align one-to-one across deformations.


\paragraph{Furthest Point Sampling}



We employ Furthest Point Sampling (FPS), similar to KeypointDeformer~\cite{jakab2021keypointdeformer, NEURIPS2024_582e9771}. To encourage robust spatial coverage, we generate a denser set of 20 reference anchors using Furthest Point Sampling (FPS) while predicting only 10 keypoints. The oversampled FPS set provides broader coverage of the object and ensures that the predicted keypoints are matched against a spatially diverse proxy. The loss minimizes the Chamfer Distance between the predicted keypoints $K$ and the FPS-selected keypoints $K_{\text{fps}}$, denoted as $\mathcal{L}_{\text{FPS}}$


\paragraph{KL Divergence for Latent Regularization}

To regularize the auxiliary latent code $z_{\text{aux}}$, we model it as a Gaussian distribution with mean $\mu$ and log-variance $\log \sigma^2$, produced by the encoder. We then apply a Kullback–Leibler (KL) divergence between the approximate posterior $q(z_{\text{aux}} \mid S_0) = \mathcal{N}(\mu, \sigma^2)$ and a standard normal prior $p(z_{\text{aux}}) = \mathcal{N}(0, I)$, denoted as $\mathcal{L}_{KL}$


This regularization encourages the auxiliary latent distribution to remain close to a unit Gaussian, promoting smoothness and generalization in the latent space. To prevent early collapse and allow the model to learn meaningful latent codes before regularization dominates, we employ a warm-up strategy we employ a warm-up strategy in which the KL weight $\lambda_{4}$ increases linearly with training step $t$, defined as $\lambda_{4} = \min\left(1.0, \frac{t}{T_{\text{warmup}}}\right)$.


\paragraph{Overall Loss Function}

The total training objective combines shape reconstruction and keypoint consistency:

\begin{equation}
    \mathcal{L} = \lambda_0 \mathcal{L}_{\text{FPS}} + \lambda_1 \mathcal{L}_{\text{diff}} + \lambda_2 \mathcal{L}_{\text{chamfer}} + \lambda_3 \mathcal{L}_{\text{mse}} + \lambda_4 \mathcal{L}_{\text{KL}}.
\end{equation}

During the initial training phase (for the first $n_{\text{init}}$ epochs), we set $\lambda_0 = \lambda_2 = \lambda_3 = 1$ and $\lambda_1 = 3$. This encourages the model to jointly learn structured keypoints and coarse shape reconstruction, while gradually introducing KL regularization. 

The KL weight $\lambda_4$ continues to follow its scheduled progression. Under this configuration, our overall loss function closely resembles the commonly used Evidence Lower Bound (ELBO) loss, which is widely adopted in variational inference frameworks~\cite{kingma2013auto}. A detailed explanation of this connection is provided in Appendix~\ref{app:elbo}.


Co-training VAEs and diffusion models have suffered from the latent space collapsing~\cite{leng2025repa}, which reduces the generation quality. The most common approach to combine VAE and diffusion models, such as Latent Diffusion Models~\cite{rombach2022high}, circumvents the latent space collapse by employing a two-stage training scheme: a VAE is first trained independently, and subsequently, a diffusion model is trained on the resulting latent representations. Inspired by this strategy, we adopted a similar setup by preventing gradient flow through the latent representation $z_0$ during diffusion training. This separation stabilizes learning and ensures that the diffusion model focuses on modeling the latent distribution, rather than interfering with the VAE reconstruction process.

\section{Experiments}

We evaluate our method across two main axes: (i) the semantic and spatial consistency of the discovered keypoints across object instances, and (ii) the generative quality of shape reconstructions. 

\subsection{Setup}

We focus our experiments on the ShapeNet dataset~\cite{shapenet2015, Yi16}, we use pre-sampled open source point clouds\footnote{https://github.com/antao97/PointCloudDatasets} for all experiments. ShapeNet shares object categories with the KeypointNet~\cite{You_2020_CVPR} benchmark, though none of the tested approaches use any keypoint annotations during training. This dataset provides a wide variety of object classes for evaluation. 


Additional experiments include keypoint-based shape interpolation and ablation studies to assess the effects of key design choices, including the number of keypoints (Appendix \ref{app:num_keypoints}), loss term contributions (Appendix \ref{app:loss}), and latent space dimensionality (Appendix \ref{app:latent}).



We evaluated keypoint quality using the Dual Alignment Score (DAS) metric~\cite{shi2021skeletonmerger} and the correlation metric introduced in KeypointDeformer~\cite{jakab2021keypointdeformer}, and assessed reconstruction performance using the Chamfer Distance (CD) and Earth Mover’s Distance (EMD), while we assessed generative performance using minimum matching chamfer distance, following~\cite{luo2021diffusionprobabilisticmodels3d}. Formal definitions of all metrics are provided in Appendix~\ref{appendix:metrics}.

\subsection{Baselines}

We compare our approach against five baselines: four keypoint-based methods and one point cloud diffusion model For keypoint prediction, we include KeypointDeformer (KPD)~\cite{jakab2021keypointdeformer}, KeyGrid~\cite{NEURIPS2024_582e9771}, Self-supervised and Coherent 3D Keypoints (SC3K)~\cite{SC3K}, and Skeleton Merger (SM)~\cite{shi2021skeletonmerger}. We executed all baselines within a unified framework to enable direct comparison. However, the released implementations varied in completeness and correctness. Hyperparameters for training were used from papers as much as possible\footnote{For several baselines, the reproduced results differ substantially from the numbers originally reported. This is largely due to incomplete code or inconsistencies in released implementations. We have taken care to align our evaluations as closely as possible with the original works, but some deviations remain unavoidable.}

SC3K does not perform reconstruction and DPM~\cite{luo2021diffusionprobabilisticmodels3d} does not predict keypoints; thus these methods are only evaluated in the relevant tasks. Conversely, for methods capable of both keypoint prediction and reconstruction, we report results across all corresponding benchmarks. Finally, while KeyGrid predicts keypoints, its decoder depends on input point features and therefore cannot be used for unconditional generation, and we excluded it from that experiment.

\section{Results}

\subsection{Keypoint Consistency}

\begin{table*}[t]
\centering
\begin{adjustbox}{max width=\textwidth}
\begin{tabular}{l|ccccc|ccccc}
\toprule
\multirow{2}{*}{Category} & \multicolumn{5}{c}{Dual Alignment Score} & \multicolumn{5}{c}{Correlation $\uparrow$} \\
& KPD & KeyGrid & SC3K & SM & Ours & KPD & KeyGrid & SC3K & SM & Ours \\
\midrule
Airplane & 0.73 $\pm$ 0.012 & 0.76 $\pm$ 0.0057 & 0.64 $\pm$ 0.035 & \textbf{0.8 $\pm$ 0.018} & 0.79 $\pm$ 0.0043 & 0.9 $\pm$ 0.037 & 0.92 $\pm$ 0.0056 & 0.83 $\pm$ 0.0058 & 0.89 $\pm$ 0.0091 & \textbf{0.96 $\pm$ 0.00077} \\
Bed & 0.63 $\pm$ 0.015 & 0.64 $\pm$ 0.011 & 0.59 $\pm$ 0.075 & 0.63 $\pm$ 0.014 & \textbf{0.71 $\pm$ 0.0028} & -- & -- & -- & -- & -- \\
Bottle & 0.56 $\pm$ 0.0033 & 0.51 $\pm$ 0.038 & 0.58 $\pm$ 0.047 & 0.63 $\pm$ 0.077 & \textbf{0.66 $\pm$ 0.0066} & -- & -- & -- & -- & -- \\
Cap & 0.77 $\pm$ 0.0084 & 0.74 $\pm$ 0.0059 & 0.7 $\pm$ 0.088 & 0.85 $\pm$ 0.064 & \textbf{0.88 $\pm$ 0.013} & 0.78 $\pm$ 0.25 & \textbf{1 $\pm$ 0} & 0.97 $\pm$ 0.023 & 0.97 $\pm$ 0.027 & 0.98 $\pm$ 0.0033 \\
Car & 0.77 $\pm$ 0.0085 & 0.73 $\pm$ 0.021 & 0.72 $\pm$ 0.094 & 0.79 $\pm$ 0.021 & \textbf{0.84 $\pm$ 0.0043} & 0.91 $\pm$ 0.047 & 0.93 $\pm$ 0.015 & 0.95 $\pm$ 0.021 & 0.94 $\pm$ 0.016 & \textbf{0.96 $\pm$ 0.01} \\
Chair & 0.83 $\pm$ 0.0056 & 0.75 $\pm$ 0.011 & 0.61 $\pm$ 0.04 & 0.79 $\pm$ 0.039 & \textbf{0.86 $\pm$ 0.00088} & 0.86 $\pm$ 0.072 & 0.9 $\pm$ 0.024 & 0.85 $\pm$ 0.016 & 0.89 $\pm$ 0.031 & \textbf{0.95 $\pm$ 0.0089} \\
Guitar & 0.56 $\pm$ 0.011 & 0.67 $\pm$ 0.0075 & 0.64 $\pm$ 0.045 & 0.63 $\pm$ 0.086 & \textbf{0.75 $\pm$ 0.0031} & 0.85 $\pm$ 0.18 & 0.98 $\pm$ 0.008 & 0.91 $\pm$ 0.01 & 0.96 $\pm$ 0.036 & \textbf{1 $\pm$ 0.00054} \\
Helmet & \textbf{0.8 $\pm$ 0.013} & 0.72 $\pm$ 0.045 & 0.59 $\pm$ 0.035 & 0.72 $\pm$ 0.018 & 0.77 $\pm$ 0.006 & -- & -- & -- & -- & -- \\
Knife & 0.59 $\pm$ 0.0081 & \textbf{0.76 $\pm$ 0.04} & 0.58 $\pm$ 0.023 & 0.63 $\pm$ 0.026 & 0.64 $\pm$ 0.00073 & 0.89 $\pm$ 0.095 & 0.95 $\pm$ 0.02 & 0.94 $\pm$ 0.0028 & 0.95 $\pm$ 0.02 & \textbf{0.97 $\pm$ 0.00032} \\
Motorbike & 0.66 $\pm$ 0.0032 & 0.6 $\pm$ 0.036 & 0.64 $\pm$ 0.057 & 0.65 $\pm$ 0.043 & \textbf{0.82 $\pm$ 0.0004} & 0.84 $\pm$ 0.041 & 0.83 $\pm$ 0.015 & 0.83 $\pm$ 0.0075 & 0.95 $\pm$ 0.043 & \textbf{1 $\pm$ 0} \\
Mug & 0.72 $\pm$ 0.011 & 0.68 $\pm$ 0.022 & 0.67 $\pm$ 0.028 & 0.77 $\pm$ 0.021 & \textbf{0.86 $\pm$ 0.02} & 0.79 $\pm$ 0.26 & 0.96 $\pm$ 0.032 & 0.94 $\pm$ 0.021 & 0.97 $\pm$ 0.039 & \textbf{1 $\pm$ 0} \\
Table & 0.86 $\pm$ 0.00098 & 0.83 $\pm$ 0.0059 & 0.7 $\pm$ 0.046 & 0.85 $\pm$ 0.0051 & \textbf{0.88 $\pm$ 0.003} & 0.89 $\pm$ 0.078 & 0.95 $\pm$ 0.0065 & 0.93 $\pm$ 0.016 & 0.98 $\pm$ 0.0056 & \textbf{0.99 $\pm$ 0.00067} \\
Vessel & 0.38 $\pm$ 0.0017 & 0.43 $\pm$ 0.047 & \textbf{0.46 $\pm$ 0.1} & 0.43 $\pm$ 0.0081 & 0.43 $\pm$ 0.0025 & -- & -- & -- & -- & -- \\
\midrule
\textbf{Average $\pm$ Std. Dev.} & 0.68 $\pm$ 0.13 & 0.67 $\pm$ 0.11 & 0.63 $\pm$ 0.058 & 0.71 $\pm$ 0.13 & \textbf{0.76 $\pm$ 0.12} & 0.69 $\pm$ 0.16 & 0.92 $\pm$ 0.047 & 0.9 $\pm$ 0.04 & 0.91 $\pm$ 0.031 & \textbf{0.98 $\pm$ 0.017} \\
\bottomrule
\end{tabular}
\end{adjustbox}
\caption{Dual Alignment Score and Correlation metrics across all algorithms. All models are trained three times, and the mean and standard deviation of the results are shown in this table. Ground truth semantic annotations are not available for all classes, therefore, not all correlation results can be calculated.}
\label{tab:keypointmetrics}
\end{table*}

\begin{figure}[htbp]
    \centering
        \includegraphics[height=0.22\linewidth]{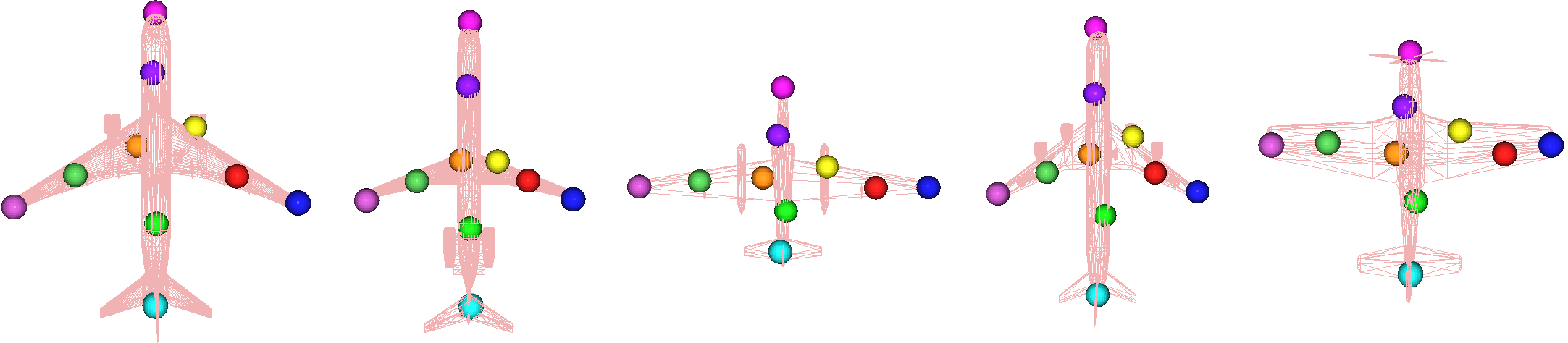}
        \includegraphics[height=0.42\linewidth]{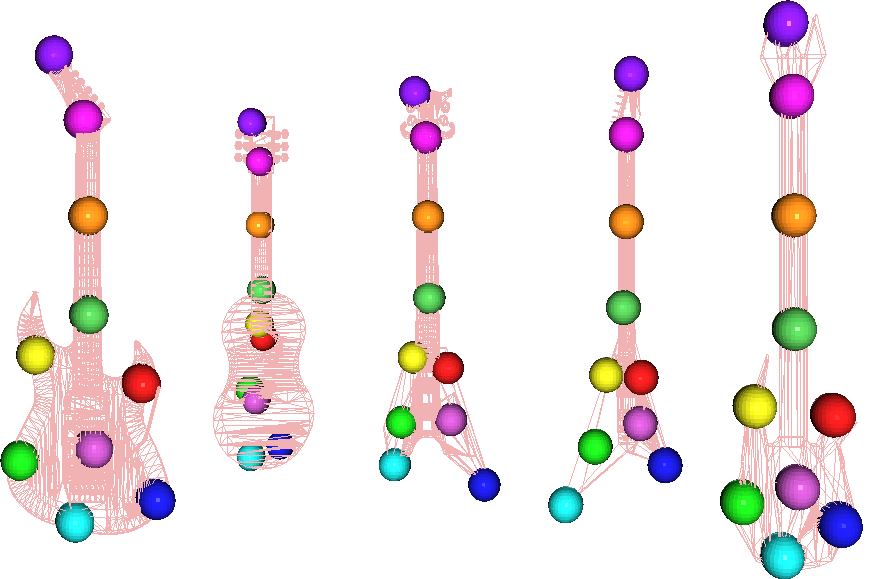}\newline
        \includegraphics[width=\linewidth]{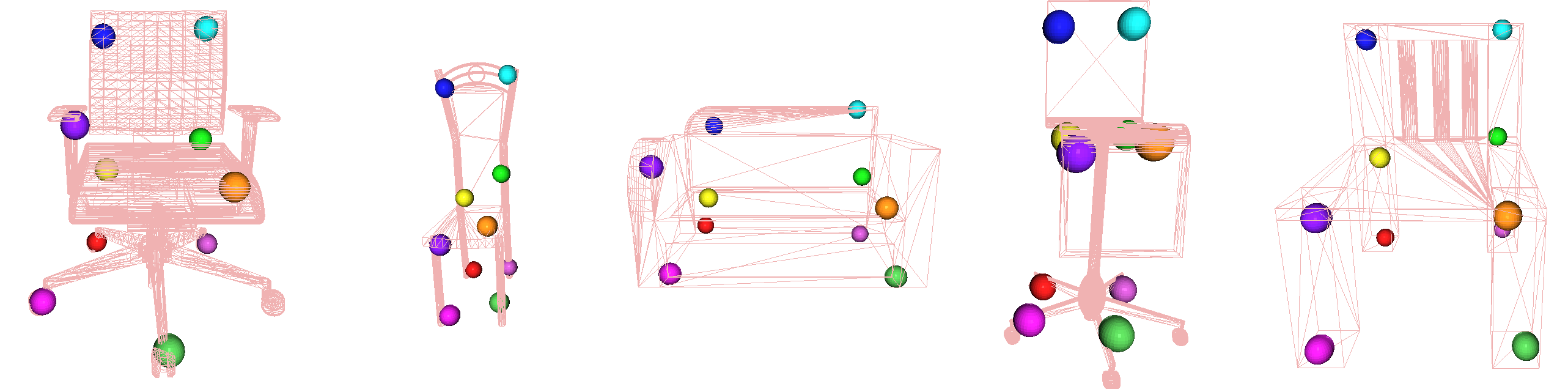}
        
    \caption{We visualize the keypoints identified by our method across different instances of the airplane (top), guitar (middle), and chair (bottom) classes, where the same keypoint ID across different instances is visualized with the same color. Keypoints predicted by our method are structurally consistent and repeatable across diverse geometries, demonstrating robustness to shape variation. }
    \label{fig:8images}
\end{figure}

As shown in Table~\ref{tab:keypointmetrics}, our method demonstrates higher keypoint consistency across object categories than prior approaches\footnote{We invested substantial effort to reproduce baselines; however, certain implementations lacked key details or exhibited inconsistencies, preventing exact replication.}. This is reflected in improved correspondence scores, indicating that our keypoints are more repeatable and stable across different instances within the same class. This advantage holds across the varied semantic complexity of the categories, demonstrating that our formulation is effective both on classes with rich structural variability and those with fewer semantic parts.

It is worth noting that keypoint correlation metrics are sensitive to the number of semantic part labels in each category. For instance, skateboard, guitar, and knife categories consist of only two part labels, making repeatable part prediction inherently easier and leading to artificially inflated scores. Additionally, class imbalance plays a role: the mug class contains just over 3\% of the data compared to the largest class (table), and the cap class only 1\%. In both of these low-data cases, KeypointDeformer~\cite{jakab2021keypointdeformer} performs poorly, suggesting that it requires substantially more training data or computational budget to generalize well\footnote{All baseline methods were trained using hyperparameters reported in their respective papers.}.

\subsection{Shape Generation and Reconstruction}

\begin{table}[t]
\centering
\begin{adjustbox}{max width=\linewidth}
\begin{tabular}{l|ccc}
\toprule
\multirow{2}{*}{Category} & \multicolumn{3}{c}{Minimum Matching Distance-Chamer Distance ($\downarrow$)} \\
& DPM & KPD & Ours \\
\midrule
Airplane & 0.0075 $\pm$ 0.0014 & 0.0089 $\pm$ 7.3e-05 & \textbf{0.0074 $\pm$ 0.00039} \\
Bed & 0.043 $\pm$ 0.0024 & 0.051 $\pm$ 0.0061 & \textbf{0.034 $\pm$ 0.00037} \\
Bottle & 0.035 $\pm$ 0.0026 & 0.009 $\pm$ 0.0015 & \textbf{0.0045 $\pm$ 0.00083} \\
Cap & 0.077 $\pm$ 0.015 & \textbf{0.018 $\pm$ 0.00093} & 0.068 $\pm$ 0.021 \\
Car & 0.0047 $\pm$ 0.00045 & \textbf{0.0041 $\pm$ 7.5e-05} & 0.0042 $\pm$ 0.00011 \\
Chair & 0.021 $\pm$ 0.0075 & 0.025 $\pm$ 0.00053 & \textbf{0.018 $\pm$ 0.00078} \\
Guitar & 0.0062 $\pm$ 0.0028 & 0.0024 $\pm$ 2.6e-05 & \textbf{0.0023 $\pm$ 0.00044} \\
Helmet & 0.033 $\pm$ 0.0022 & 0.035 $\pm$ 0.0044 & \textbf{0.02 $\pm$ 0.0017} \\
Knife & 0.033 $\pm$ 0.022 & \textbf{0.0037 $\pm$ 0.00023} & 0.0078 $\pm$ 0.0018 \\
Motorbike & 0.054 $\pm$ 0.012 & \textbf{0.0067 $\pm$ 0.0004} & 0.008 $\pm$ 0.0011 \\
Mug & 0.044 $\pm$ 0.00073 & 0.02 $\pm$ 0.0012 & \textbf{0.01 $\pm$ 0.00043} \\
Table & \textbf{0.013 $\pm$ 0.00076} & 0.029 $\pm$ 0.00035 & 0.017 $\pm$ 0.00074 \\
Vessel & \textbf{0.0075 $\pm$ 0.0014} & 0.01 $\pm$ 0.00024 & 0.0077 $\pm$ 0.00031 \\
\midrule
\textbf{Average $\pm$ Std. Dev.} & 0.029 $\pm$ 0.021 & 0.017 $\pm$ 0.014 & \textbf{0.016 $\pm$ 0.017} \\
\bottomrule
\end{tabular}
\end{adjustbox}
\caption{The minimum matching distance between generated objects and the testing set. Our approach outperforms other approaches without relying on external references.}
\label{tab:multi_metrics_mean_std}
\end{table}

\begin{table*}[t]
\centering
\begin{adjustbox}{max width=\textwidth}
\begin{tabular}{l|ccccc}
\toprule
\textbf{Metric} & DPM & KPD & KeyGrid & SM & Ours \\
\midrule
\textbf{Inputs to Decoder} 
& 35 & 30 + Reference Mesh & 6671 & 16414 & 35  \\
\midrule
\textbf{Chamfer Distance ($\downarrow$)} 
& 0.042 $\pm$ 0.031 
& 0.049 $\pm$ 0.038 
& \textbf{0.0065 $\pm$ 0.0066} 
& 0.015 $\pm$ 0.012 
& 0.021 $\pm$ 0.019 \\
\textbf{Earth Movers Distance ($\downarrow$)} 
& 4.9e-05 $\pm$ 2.3e-05 
& 4.6e-05 $\pm$ 2.4e-05 
& \textbf{3.1e-05 $\pm$ 1.9e-05} 
& 4.3e-05 $\pm$ 1.3e-05 
& 5.7e-05 $\pm$ 2.7e-05 \\
\bottomrule
\end{tabular}
\end{adjustbox}
\caption{Reconstruction-based results on ShapeNet. KeyGrid has strong performance but relies on the input point cloud to complete reconstruction}
\label{tab:reconstruct}
\end{table*}

Both our method and KeypointDeformer are capable of generating new shapes from keypoints rather than only reconstructing input ones. Our method outperforms KeypointDeformer on 9 out of 13 classes, demonstrating strong performance on the ability to synthesize new objects. However, KeypointDeformer relies on deforming a reference mesh toward target keypoints, making it less effective when facing substantial intra-class shape diversity, as generation quality depends heavily on the chosen reference mesh. This deformation-based strategy can therefore produce unrealistic shapes when the reference is poorly aligned with the sampled set of keypoints. Our approach generates point clouds directly, without relying on a reference mesh, enabling more flexible and diverse shape synthesis. Unlike deformation-based methods, which inherit structural smoothness from a template, our fully generative formulation learns geometry directly from data. This can introduce surface noise in the synthesized shapes, but in exchange enables a significantly higher generative flexibility.

DPM~\cite{luo2021diffusionprobabilisticmodels3d} originally employs a 256-dimensional latent code. For a fairer comparison, we reduce this dimensionality to 35 to match our setup. This results in a notable drop in reconstruction quality relative to the original implementation, suggesting that the model relies heavily on a high-capacity latent space to retain geometric structure. These findings indicate that diffusion models without explicit spatial priors may struggle under tight latent bottlenecks, highlighting the importance of structural encoding when operating with compact latent representations.

We also evaluate reconstruction quality using Chamfer Distance and Earth Mover’s Distance, as reported in Table~\ref{tab:reconstruct}. However, the pure reconstruction setting has limited practical relevance, as most existing methods depend directly on input features, which inflates their reconstruction performance but does not translate to general utility. KeyGrid achieves strong reconstruction performance, likely due to its architectural design, where the decoder receives early-layer feature information from the encoder in addition to the latent inputs, simplifying the reconstruction task. SkeletonMerger also conditions directly on the input shape when creating its skeleton activation, restricting its usefulness for unconditional generation. In contrast, our decoder does not access any encoder features, making the reconstruction task inherently more challenging; consequently, our reconstruction performance is weaker in this setting.




\subsection{Keypoint Interpolation}

\begin{figure}[htbp]
    \centering
        \includegraphics[trim=5cm 1cm 5cm 0cm, clip, width=0.12\textwidth]{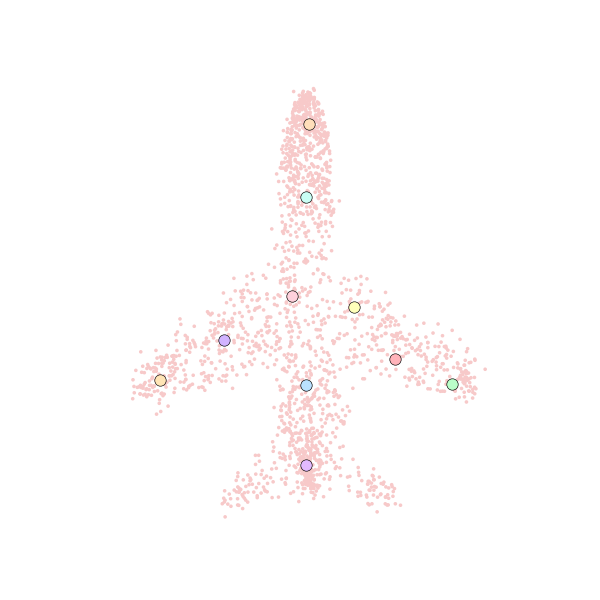}\hfill
        \includegraphics[trim=5cm 1cm 5cm 0cm, clip, width=0.12\textwidth]{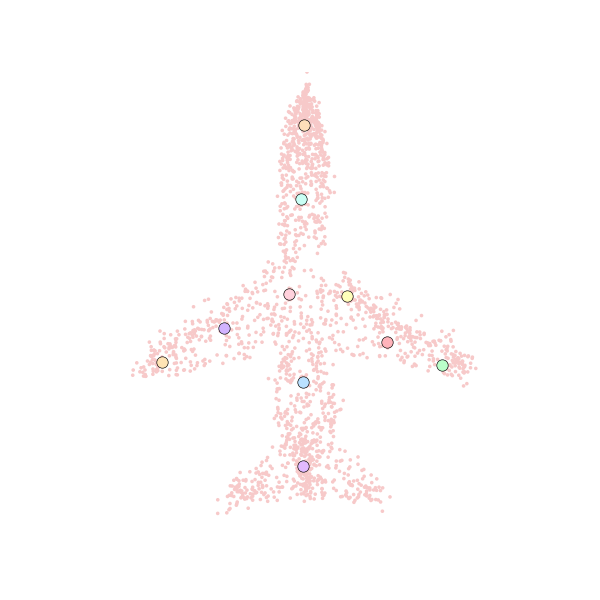}\hfill
        \includegraphics[trim=5cm 1cm 5cm 0cm, clip, width=0.12\textwidth]{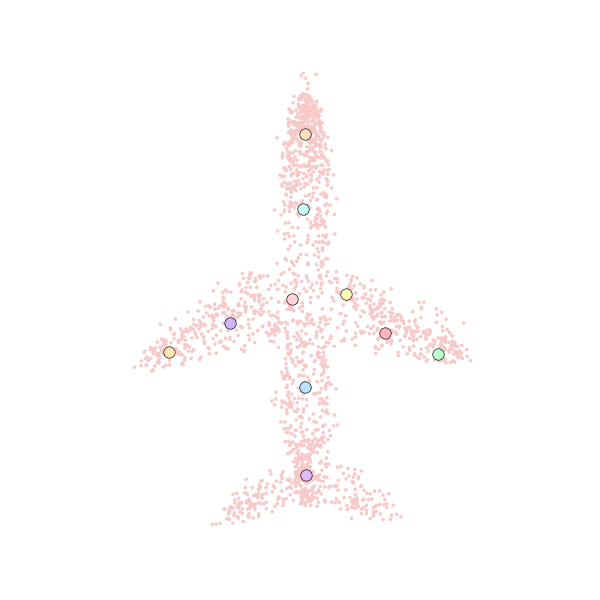}\hfill
        \includegraphics[trim=5cm 1cm 5cm 0cm, clip, width=0.12\textwidth]{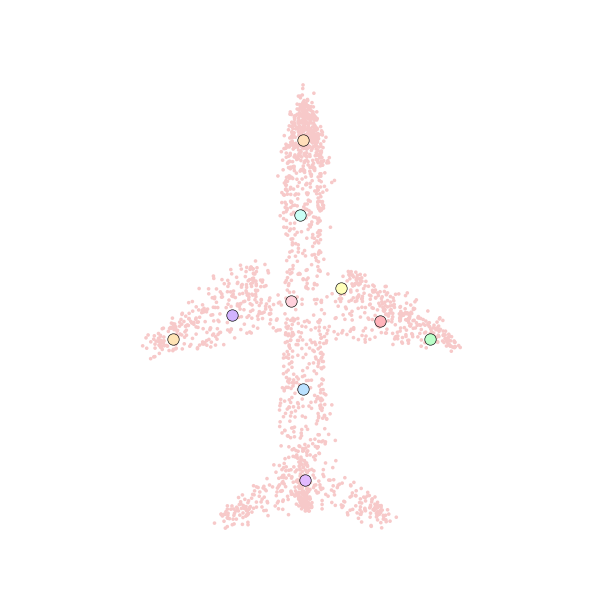}\hfill
        \includegraphics[trim=5cm 1cm 5cm 0cm, clip, width=0.12\textwidth]{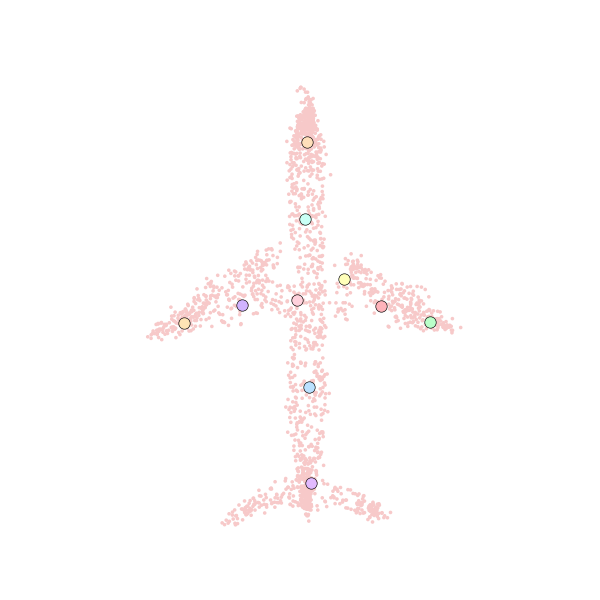}\hfill
        \includegraphics[trim=5cm 1cm 5cm 0cm, clip, width=0.12\textwidth]{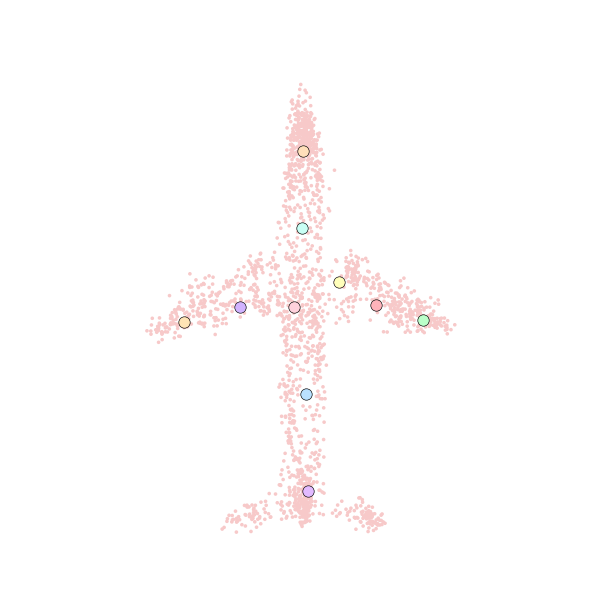}\hfill
    \caption{Linear interpolation in the learned keypoint space between two airplane shapes. The top-left and bottom-right point clouds are reconstructions of point clouds from the test set samples from the dataset, while the four intermediate shapes are generated by decoding linearly interpolated keypoints. The smooth transitions demonstrate the continuity and semantic structure of the learned representation.}
    \label{fig:interpolation}
\end{figure}

To investigate the structure and continuity of the learned keypoint representation, we performed linear interpolation between the keypoints of two shapes from the same category. Specifically, given two input point clouds, we encode each into its corresponding set of keypoints, then interpolate linearly between the keypoint sets. Each interpolated keypoint configuration is decoded back into a point cloud using our generative model. As shown in Figure~\ref{fig:interpolation}, the decoded shapes evolve continuously and maintain realistic, coherent structure throughout the interpolation, indicating that the generative model captures a smooth and expressive representation that generalizes well to unseen keypoint configurations.

\section{Limitations}
\label{sec:limitations}
\begin{figure}
    \centering
  \includegraphics[width=0.45\linewidth]{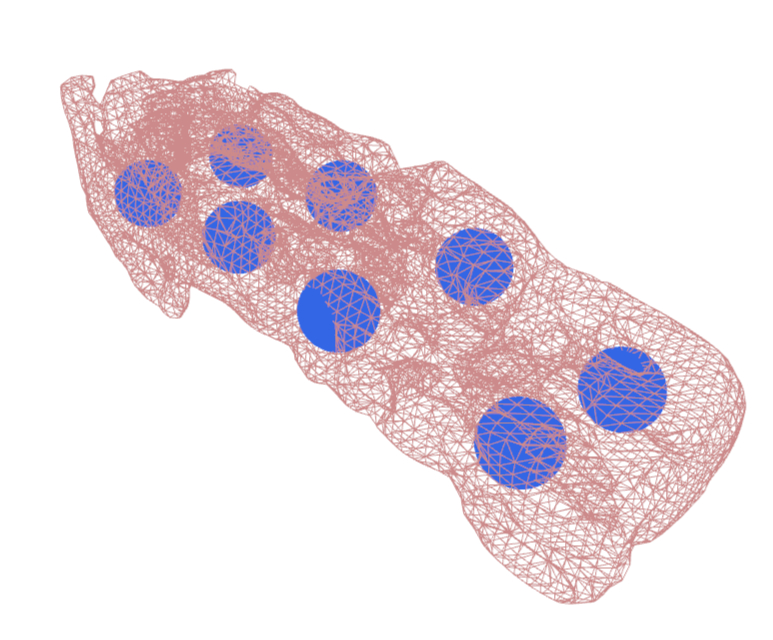}
  \caption{The result of using NKSR~\cite{huang2023nksr}, a deep-learning based algorithm to predict the mesh from a noisy point cloud, such as the one calculated from our diffusion model. The resulting mesh is noisy and not realistic.}
  \label{fig:failure}
\end{figure}

  While the proposed method demonstrates strong performance on synthetic datasets, several limitations constrain its broader applicability. First, the model produces 3D point clouds instead of meshes, and the point sets are noticeably sparse, primarily due to the 2,048 point sampling used during training. Increasing the density of the training point clouds could mitigate this limitation. The output point clouds tend to still contain noise, particularly in regions with limited training coverage, which further challenges high-quality mesh reconstruction.
  As a result of the sparsity and noise, reconstructing high-quality watertight meshes remains a nontrivial post-processing challenge. For example, applying a learning-based meshing method like NKSR~\cite{huang2023nksr} to outputs from our diffusion model results in meshes that are noisy and unrealistic, as shown in Figure~\ref{fig:failure}. A promising direction for future work is to extend the diffusion process to operate directly on mesh representations, as explored in recent work like PolyDiff~\cite{alliegro2023polydiffgenerating3dpolygonal}, which could improve surface fidelity and eliminate the need for point-based postprocessing. Second, the approach assumes that input shapes used for training are prealigned and consistently scaled. In real-world scenarios, achieving such normalization is difficult and often requires manual intervention or additional pre-processing steps. Third, the method does not explicitly leverage inherent geometric properties of objects such as symmetry or part relationships. Incorporating these priors could improve both the consistency of learned keypoints and the quality of generated shapes.

{
    \small
    \bibliographystyle{ieeenat_fullname}
    \bibliography{refs}
}

\clearpage
\setcounter{page}{1}
\maketitlesupplementary

\appendix

\section{Soft Projection Mechanism}
\label{app:softproject}

To ensure that predicted keypoints remain close to the underlying surface geometry, we employ a \emph{soft projection} operator. Given a set of predicted keypoints and a dense set of surface points, the operator computes a differentiable approximation of nearest-neighbor projection by using a softmin over Euclidean distances.

Given a predicted keypoint set $\hat{X} = \{\hat{\mathbf{x}}_k\}_{k=1}^K$ and a set of surface points
$S = \{\mathbf{s}_i\}_{i=1}^N$, the goal of the soft projection operator is to map each predicted point smoothly toward the surface. For each predicted point $\hat{\mathbf{x}}_k$ we compute its distances to all surface points:
\begin{equation}    
d_{k,i} = \|\hat{\mathbf{x}}_k - \mathbf{s}_i\|_2.
\end{equation}

These distances are transformed into a normalized weight distribution using a softmin with temperature $\tau$:
\begin{equation}    
w_{k,i} = \frac{\exp\!\left(-d_{k,i}/\tau\right)}
     {\sum_{j=1}^N \exp\!\left(-d_{k,j}/\tau\right)}.
\end{equation}

The soft-projected point is then given by the weighted average
\begin{equation}    
\tilde{\mathbf{x}}_k = \sum_{i=1}^N w_{k,i}\,\mathbf{s}_i.
\end{equation}

As $\tau \to 0$, the distribution $w_{k,i}$ approaches a one-hot vector centered at the true nearest neighbor, recovering the standard nearest--neighbor projection. For larger $\tau$, the
projection becomes smoother and incorporates information from a neighborhood of surface points.

\section{Shape Diffusion Process}
\label{app:edm_forward}

The forward diffusion process gradually transforms a clean shape into pure noise by adding Gaussian noise over a series of diffusion steps. This diffusion process generates various noisy shape samples for the model to learn how to reverse them during training, enabling shape generation from noise.

In contrast to standard Denoising Diffusion Probabilistic Models (DDPMs), EDM defines the forward noising process via a continuous noise scale $\sigma_t \in [\sigma_{\min}, \sigma_{\max}]$, where noise is added directly as:

\begin{equation}
q(S_{t} \mid S_0) = \mathcal{N}(S_t; S_0, \sigma_t^2 I)
\end{equation}

Here, $S_0 \in \mathbb{R}^{N \times 3}$ is the clean point cloud, and $S_t$ is its noisy counterpart at noise level $\sigma_t$. This forward process admits a reparameterization, allowing us to sample $S_t$ as a deterministic function of $S_0$, $\sigma_t$, and Gaussian noise $\epsilon \sim \mathcal{N}(0, I)$:

\begin{align}
    S_t &= S_0 + \sigma_t \epsilon
\end{align}
Furthermore, $\sigma_T = \sigma_{\text{max}}$ is chosen to be sufficiently large such that the distribution $S_T$ closely approximates a Gaussian, i.e., $S_T \sim \mathcal{N}(0, \sigma_T^2 I)$. We sample noise levels $\sigma_t$ are sampled from a log-Gaussian distribution which is customized to heavily sample on the mid-level noise that is critical for generating samples~\cite{karras2022elucidating}

\section{Shape Reconstruction using EDM}
\label{app:shape_edm}

The reverse process reconstructs clean 3D shapes from Gaussian noise through iterative denoising steps governed by the EDM framework. While EDM typically operates unconditionally, we condition the denoising process on the vector $z_0$, which aims to serve as a structural prior. 

The model is trained to predict the cleaned point cloud from $S_t$ using a preconditioned score network $D_\theta$. To stabilize training, we adopt EDM's formulation that normalizes both input and output to unit variance, resulting in the following scaling coefficients:

\begin{align}
\begin{cases}
    c_\text{in} &= \frac{1}{\sqrt{\sigma^2 + \sigma_\text{data}^2}} \\
    c_\text{out} &= \frac{\sigma \cdot \sigma_\text{data}}{\sqrt{\sigma^2 + \sigma_\text{data}^2}} \\
    c_\text{skip} &= \frac{\sigma_\text{data}^2}{\sigma^2 + \sigma_\text{data}^2} \\
    c_\text{noise} &= \frac{1}{4} \log \sigma
\end{cases}
\end{align}

The final denoised prediction is then given by:

\begin{equation}
D_\theta(S_\sigma, \sigma, z_0) = c_\text{skip} \cdot S_\sigma + c_\text{out} \cdot F_\theta(c_\text{in} \cdot S_\sigma, c_\text{noise}, z_0)
\end{equation}

where $F_\theta$ is the neural network backbone conditioned on the latent context vector $z_0$. The parameter $\sigma_\text{data}$ is a fixed hyperparameter that estimates the dataset’s average standard deviation; we set $\sigma_\text{data} = 0.3$ for all experiments.

\paragraph{The weighting function} is defined as:
\begin{equation}
w(\sigma_t) = \left( \sigma_t^2 + \sigma_\text{data}^2 \right)^{-2},
\end{equation} following~\cite{karras2022elucidating}, to balance contributions from different noise levels and emphasize mid-level noise samples during training.

\section{Derivation of ELBO Loss}
\label{app:elbo_loss_derive}

The objective is to maximize the probability of point clouds $S_0$. And we define a latent representation $z_0$ that determines the point cloud, these could be the keypoint, object type, and the shape-specific feature and latent representation of the point cloud $S_{1:T}$ that iteratively shape the original point cloud from complete noise $S_{T}$ to $S_0$.
{\footnotesize
\begin{align}
&\log p(S_0) &\notag \\
& \geq \mathbb{E}_{q(z_0, S_{1:T}|S_0)} \left[ \log \frac{p(s_{0:T}, z_0)}{q(z_0, S_{1:T}|S_0)} \right] \\
    &= \mathbb{E}_{q(z_0, S_{1:T}|S_0)} \left[ \log \frac{p(s_T)p_\phi(z_0)\prod_{t=0}^{T-1} p(S_{t}|S_{t+1}, z_0)}{q(z_0, S_{1:T}|S_0)} \right] \\
    &= \mathbb{E}_{q(z_0, S_{1:T}|S_0)} \log p(S_0|S_1, z_0) + \mathbb{E}_{q(z_0|S_0)} \left[\frac{p(z_0)}{q(z_0|S_0)} \right] \notag \\
    &\quad + \sum_{t=1}^{T-1} \mathbb{E}_{q(z_0, S_{1:T}|S_0)} \left[ \log \frac{p(S_t|S_{t+1}, z_0)}{q(S_{t}|S_{t-1})} \right]
\end{align}
}
\section{Connection to ELBO Loss}
\label{app:elbo}

Our model formulation has a strong resemblance to latent diffusion models.  Given a point cloud, we wish to find a latent representation $z_0$ of the point cloud that contains the necessary information about the point cloud. 

Given the latent representation $z_0$, we want to reconstruct the original point cloud $S_0$ using a diffusion decoder. Similar to the Variational Diffusion Model~\cite{kingma2021variational}, the objective is to maximize the probability of generating point clouds $S_0$. 

{\footnotesize
\begin{align}
    &\log p(S_0) \notag \\
    &\geq \mathbb{E}_{q(z_0, S_{1:T}|S_0)} \left[ \log \frac{p_\theta(s_{0:T}, z_0)}{q(z_0, S_{1:T}|S_0)} \right] \\
    &= \mathbb{E}_{q(z_0, S_{1:T}|S_0)} \log p(S_0|S_1, z_0) + \mathbb{E}_{q(z_0|S_0)} \left[\frac{p_\phi(z_0)}{q_\phi(z_0|S_0)} \right] \\
    &\quad + \sum_{t=1}^{T-1} \mathbb{E}_{q(z_0, S_{1:T}|S_0)} \left[ \log \frac{p_\theta(S_t|S_{t+1}, z_0)}{q(S_{t}|S_{t-1})} \right] \label{fig:elbo-derivation}
\end{align}
}
where we assume $p(S_T) = q(S_T)$ which is the normal distribution.

The evidence lower bound (ELBO) provides a principled training objective for models that combine latent variable inference and generative modeling, such as ours. In our model, training is closely tied to maximizing the ELBO on the data log-likelihood $\log P(S_0)$. The derivation shown in Equation~\ref{fig:elbo-derivation} decomposes $\log P(S_0)$ into three key terms:

\begin{enumerate}
    \item \textbf{Reconstruction Term:}

    $\mathbb{E}_{q(z_0 \mid S_0)} \left[ \log P(S_0 \mid S_1, z_0) \right]$,  
    which encourages the latent variable $z_0$ (i.e., keypoints and auxiliary features) to capture sufficient information to reconstruct the original shape.

    \item \textbf{Diffusion Matching Term:}

    $\mathbb{E}_{q(z_0, S_{1:T}|S_0)} \left[ \log \frac{p_\theta(S_t|S_{t+1}, z_0)}{q(S_{t}|S_{t-1})} \right]$,  
    which ensures that the learned denoising process can accurately invert the forward noising process, conditioned on the keypoints.

    \item \textbf{Latent Regularization Term:}

    $\mathbb{E}_{q(z_0 \mid S_0)} \left[ \log \frac{P(z_0)}{q(z_0 \mid S_0)} \right] = -\text{KL}(q(z_0 \mid S_0) \,||\, P(z_0))$,  
    which regularizes the latent space, encouraging the auxiliary latent codes to follow a unit Gaussian prior. 
\end{enumerate}

The latent regularization term in VAE is usually a Gaussian prior.  Our model aims to learn unsupervised keypoint detection and therefore, instead of a noninformative Gaussian prior over the latent representation, we use keypoint-specific loss functions and only enforce the Gaussian latent regularization on the $z_{\text{aux}}$ space.

Each of these components corresponds to specific terms in our final training loss:
\begin{itemize}
    \item The \textbf{diffusion reconstruction loss} $\mathcal{L}_{\text{diff}}$ captures the reconstruction and denoising components, encouraging the model to recover clean shapes from noisy inputs conditioned on keypoints.
    \item The \textbf{KL divergence loss} $\mathcal{L}_{\text{KL}}$ explicitly regularizes the auxiliary latent code toward a standard Gaussian prior.
    \item The \textbf{Chamfer loss} $\mathcal{L}_{\text{chamfer}}$, \textbf{deformation consistency loss} $\mathcal{L}_{\text{mse}}$, and \textbf{furthest point sampling loss} $\mathcal{L}_{\text{FPS}}$ further regularize the structured part of the latent space, promoting spatial and semantic consistency of the keypoints.
\end{itemize}

Thus, our full objective aligns with maximizing the ELBO while incorporating additional geometric regularization to improve the interpretability and robustness of the learned keypoints.

\section{Model}
\label{app:model}

Our model is structured as an autoencoder composed of a PointTransformerV3-based encoder and a diffusion-based decoder.

\paragraph{Encoder.}  


The encoder maps an input 3D point cloud to a structured latent representation in terms of keypoints. As a backbone, we use PointTransformerV3~\citep{wu2024pointtransformerv3simpler}, a Transformer-based architecture designed for point cloud understanding. Given a batched set of point clouds, PointTransformerV3 operates directly on raw 3D coordinates and produces per-point features $\mathbf{f}_i \in \mathbb{R}^{64}$. To preserve spatial information and enrich the geometric signal, we augment these features with a Fourier positional encoding~\cite{li2021learnablefourierfeaturesmultidimensional} of the coordinates. Concretely, we sample fixed random frequencies $\mathbf{F} \in \mathbb{R}^{3 \times F}$ and define
\begin{equation}    
\gamma(\mathbf{x}_i)
=
\big[
\sin\left(2\pi \mathbf{x}_i \mathbf{F}\right),\;
\cos\left(2\pi \mathbf{x}_i \mathbf{F}\right)
\big]
\in \mathbb{R}^{2F}.
\end{equation}

For each point, we concatenate its backbone feature and positional encoding, $[\mathbf{f}_i, \gamma(\mathbf{x}_i)]$, and project them to an embedding dimension $D$ via a linear layer.

On top of these point embeddings, we introduce $K$ learnable query vectors $\{\mathbf{q}_k\}_{k=1}^K$, with $\mathbf{q}_k \in \mathbb{R}^D$. A multi-head cross-attention layer then lets these queries attend to the set of point embeddings, yielding attention weights $a_{k,i}$ over points. Each keypoint is defined as a convex combination of the input coordinates,
\[
\hat{\mathbf{x}}_k = \sum_i a_{k,i}\,\mathbf{x}_i,
\]
so that keypoints always lie inside the convex hull of the observed point cloud. In addition, the attended query outputs are pooled and passed through a small MLP to produce the auxillary code ($z_{aux}$).

\paragraph{Decoder.}  

The decoder is a point-wise denoising network that predicts cleaned coordinates from noisy point clouds. It is conditioned on two signals: the diffusion timestep and the shape-level latent code produced by the encoder.

The timestep $\beta$ is first mapped to a sinusoidal embedding using a standard positional encoding. This embedding is then processed through an MLP to produce a time-dependent conditioning vector. In parallel, the latent code from the encoder is projected through its own MLP. These two vectors are concatenated to form a single global conditioning representation, which is broadcast to every point in the cloud.

The network begins by applying a cross-attention block, where each point uses its coordinates as queries and attends to the global conditioning context. This provides an initial geometry-aware correction to the noisy points.

Following this, the decoder applies a series of FiLM-modulated~\citep{perez2018film} MLP layers. Each FiLM layer receives both the per-point features and the global conditioning vector, and adjusts its activations accordingly. These layers operate independently across points, allowing the model to adapt its behavior based on both the noise level and the global structural information.

To encourage global consistency, the intermediate point-wise features are pooled across all points to produce a global summary, which is then concatenated back to each point’s local features. This fused representation is passed through additional FiLM-modulated layers that map the combined local and global information back to predicted denoised 3D coordinates.

\section{Differentiable Deformation Process}
\label{appendix:deformations}

The deformation operator $\mathcal{T}$ used to generate transformed shapes $S_d = \mathcal{T}(S_t)$ consists of a sequence of structured, differentiable operations that simulate realistic object variations. These deformations are parameterized by a transformation matrix $M \in \mathbb{R}^{3 \times 3}$, applied to the input point cloud $S_t \in \mathbb{R}^{N \times 3}$. The components of $\mathcal{T}$ include:
    
\paragraph{Stretching}: Stretching is implemented as a scaling operation along a random unit direction $\mathbf{v} \in \mathbb{R}^3$.  
A stretch factor $\lambda \in [1, \lambda_{\max}]$ (with $\lambda_{\max} = 2.2$) is sampled, and the matrix

\begin{equation}    
M_{\text{stretch}} = I + (\lambda - 1)\, \mathbf{v}\mathbf{v}^\top
\end{equation}

is applied. This expands the point cloud along direction $\mathbf{v}$ while leaving orthogonal directions unchanged.

\paragraph{Bending} Bending is simulated using a shear transformation between two randomly chosen axes.  
For each sample, an input axis $i \in \{0,1,2\}$ and a distinct output axis $o \neq i$ are selected.  
A bending coefficient $\alpha \in [-\beta, \beta]$ (with $\beta$ the maximum bending factor, $\beta = 1.8$) is drawn, and the bending matrix is

\begin{equation}    
M_{\text{bend}} = I + \alpha\, \mathbf{E}_{o,i},
\end{equation}

where $\mathbf{E}_{o,i}$ has a single nonzero entry at $(o,i)$.  
This adds a fraction of coordinate $p_i$ to $p_o$, producing a randomized shear that approximates bending.

\paragraph{Twisting} Twisting is implemented as a rotation around the $x$-axis, where the rotation magnitude depends on the mean $x$-coordinate of the shape.  
Let $\bar{x}$ denote the mean of the $x$-coordinates in the point cloud. A twist factor $\gamma \in [0, \theta_{\max}]$ (with $\theta_{\max} = 1.9$) is sampled to give

\begin{equation}
\theta = \gamma \cdot \bar{x}.
\end{equation}

The corresponding twist matrix is

\begin{equation}    
M_{\text{twist}} =
\begin{bmatrix}
1 & 0 & 0 \\
0 & \cos\theta & -\sin\theta \\
0 & \sin\theta & \cos\theta
\end{bmatrix}.
\end{equation}

This introduces a smooth twist whose magnitude depends on the geometry of the input.

\paragraph{Tapering} Tapering is modeled as an affine transformation that makes the $x$ and $z$ coordinates depend on the $y$ coordinate.  
A taper factor $\tau \in [0, \tau_{\max}]$ (with $\tau_{\max} = 1.6$) is sampled, and the matrix

\begin{equation}
M_{\text{taper}} =
I +
\begin{bmatrix}
0 & \tau & 0 \\
0 & 0 & 0 \\
0 & \tau & 0
\end{bmatrix}
\end{equation}

is applied.  
This modifies the geometry according to

\begin{equation}
x' = x + \tau y, \qquad z' = z + \tau y.
\end{equation}

\paragraph{Rotation}

Finally, a small global rotation around the vertical ($y$) axis is applied.  
A rotation angle $\phi \in [-\phi_{\max}, \phi_{\max}]$ (with $\phi_{\max} = \frac{\pi}{6}$) is sampled, and

\[
M_{\text{rot}} =
\begin{bmatrix}
\cos\phi & 0 & \sin\phi \\
0 & 1 & 0 \\
-\sin\phi & 0 & \cos\phi
\end{bmatrix}
\]

is composed with the other transformations.

\section{Metrics}
~\label{appendix:metrics}

\paragraph{Keypoint Correlation}: To evaluate keypoint consistency, we adopt the same metric as~\cite{jakab2021keypointdeformer}. We compare predicted keypoints to point clouds segmented into each part~\cite{Yi16}. Each point cloud $\mathcal{S} \in \mathbb{R}^{N \times 4}$ contains 3D points and one associated part label per point label. For each keypoint $\mathbf{k}_i \in \mathbf{K} \in \mathbb{R}^{d \times 3}$, we calculate distances to all points in $\mathcal{S}$:

\begin{equation}
\mathbf{D}_i = \| \mathbf{k}_i - \mathbf{p}_j \|_2, \quad \forall j \in \{1, \ldots, N\}.\    
\end{equation}

Points within a threshold $\tau = 0.05$ are selected, and their labels $\mathbf{L}_i$ are used to construct a binary matrix $\mathbf{C} \in \{0, 1\}^{d \times L}$, where:

\begin{equation}
\mathbf{C}_{i, l} = 1 \iff l \in \mathbf{L}_i.
\end{equation}

We define the keypoint-part correlation matrix $\mathbf{M} \in [0, 1]^{d \times L}$ such that each entry $\mathbf{M}_{i, l}$ denotes the fraction of samples in which keypoint $i$ is associated with part label $l$. Formally:

\begin{equation}
\mathbf{M}_{i, l} = \frac{1}{S} \sum_{s=1}^{S} \mathbf{C}^{(s)}_{i, l},
\end{equation}

where $\mathbf{C}^{(s)} \in \{0, 1\}^{d \times L}$ is the binary keypoint-to-label association matrix for the $s$-th sample, and $S$ is the total number of samples. A value of $\mathbf{M}_{i,l} = 1$ indicates that keypoint $i$ consistently aligned with part $l$ across all samples.

\begin{equation}
\text{Consistency Score} = \frac{1}{L} \sum_{l=1}^{L} \max_{i} \mathbf{M}_{i, l}.
\end{equation}

This score captures how well the keypoint aligns with each semantic part.

\paragraph{Dual Alignment Score (DAS).} 

To further evaluate semantic consistency of predicted keypoints across different shapes, we adopt the Dual Alignment Score (DAS)~\cite{shi2021skeletonmerger}. The core idea is to measure whether a predicted keypoint corresponds to the same semantic part across aligned shapes, in agreement with human keypoint annotations.

Given a \emph{reference} point cloud with human-annotated keypoints and an \emph{evaluation} point cloud, DAS is computed in two directions:

\begin{enumerate}
    \item \textbf{Assign:} On the reference point cloud, each predicted keypoint is assigned a semantic label by finding its nearest human-annotated keypoint.
    \item \textbf{Infer:} Since predicted keypoints are aligned across shapes, the same semantic label is propagated to the corresponding predicted keypoint in the evaluation point cloud.
    \item \textbf{Match:} On the evaluation point cloud, we check whether the nearest human-annotated keypoint to this predicted keypoint has the same semantic label. This yields an accuracy score.
\end{enumerate}

The reverse direction is computed analogously: semantic labels are assigned from annotations to predicted keypoints, propagated to the other shape, and matched back against annotations. The final DAS is the average of accuracies from both directions:

\begin{equation}
\text{DAS} = \tfrac{1}{2} \Big( \text{Acc}(\text{pred} \rightarrow \text{anno}) + \text{Acc}(\text{anno} \rightarrow \text{pred}) \Big).
\end{equation}

A higher DAS indicates that predicted keypoints consistently correspond to the same annotated semantic keypoint across different shapes, reflecting stable and semantically meaningful keypoint predictions.

\textbf{Reconstruction Quality}

We use two metrics to assess the quality of point cloud reconstruction. 

\textbf{Chamfer Distance (CD):} For two point clouds $A, B \subset \mathbb{R}^{N \times 3}$, we calculate the symmetrical chamfer distance ($\frac{1}{2}(d_{CD}(A \rightarrow B) + d_{CD}(B \rightarrow A))$ (see Equation \ref{eqn:chamfer_distance_general}).

\textbf{Earth Mover's Distance (EMD):} Assuming $A$ and $B$ have the same number of points, EMD is the minimum cost of transforming one point cloud into the other:

\begin{equation}
d_{\text{EMD}}(A, B) = \min_{\phi: A \rightarrow B} \frac{1}{|A|} \sum_{a \in A} \|a - \phi(a)\|_2,
\end{equation}

where $\phi$ is a bijection between the two point sets.

\textbf{Distribution Quality (MMD-CD)} 

To evaluate the generative quality of our model, we compute Minimum Matching Distances using Chamfer Distance (MMD-CD) between generated and real point clouds. These metrics assess both fidelity and diversity of the generated shape distribution.

Because all decoders (apart from DPM~\cite{luo2021diffusionprobabilisticmodels3d}) are conditioned on structured latent representations, we cannot directly generate shapes by sampling from a standard Gaussian prior, as is commonly done in unstructured latent generative models. Instead, we synthesize new shapes by first sampling plausible keypoint representations. We collect all training-set keypoints, and fit a PCA model to project them into a lower-dimensional subspace. A kernel density estimator (KDE) is then fit in this subspace, from which we sample new keypoints. These samples are mapped back to the original keypoint space using the inverse PCA transform. To form complete latent codes for our model, we concatenate each sampled keypoint vector with the mean auxiliary latent $\bar{z}_{\text{aux}}$, estimated from the training set. The resulting latent vectors $z_0$ are used to condition the diffusion model, generating shapes from Gaussian noise.

The MMD between a generated set $\mathcal{G} = {G_1, \ldots, G_M}$ and a reference set $\mathcal{R} = {R_1, \ldots, R_N}$ is defined as:

\begin{equation}
\text{MMD}(\mathcal{G}, \mathcal{R}) = \frac{1}{|\mathcal{R}|} \sum_{R \in \mathcal{R}} \min_{G \in \mathcal{G}} d(R, G),
\end{equation}

where $d(\cdot, \cdot)$ denotes a pairwise distance metric between point clouds. We fix $m=n$ and always generate point cloud sets the same size as the training set.

\section{Training Details}
\label{app:training}

We train all models for 200 epochs using an NVIDIA RTX 3090, each model takes around 12 hours to train. Each training step uses a batch size of 120, which is split into 10 gradient accumulation steps due to memory constraints. We optimize using the Adam optimizer with a learning rate of $1 \times 10^{-3}$. Our generative model employs a diffusion network, which operates on 3D point clouds with conditioning from shape-level latent vectors. The input to the network consists of a point cloud $x \in \mathbb{R}^{12 \times 2048 \times 3}$ and a context vector consisting of 10 keypoints and 5 latent dimension (i.e., 35 dimensions). 

\section{Full Loss Function Definitions}
\label{app:full_loss_equations}

For completeness, we now present all losses in full mathematical form.

\paragraph{Diffusion Reconstruction Loss ($\mathcal{L}_{\text{diff}}$):}

This loss guides the denoising model to reconstruct clean shapes from noisy inputs via Chamfer Distance:

\begin{align}
\mathcal{L}_{\text{diff}}
&=
\mathbb{E}_{\sigma_t, S_0, \epsilon}
\left[
w(\sigma_t)
\cdot
\mathcal{L}_{\text{CD}}\big(D_\theta(S_0 + \sigma_t \epsilon, \sigma_t, z_0),\, S_0\big)
\right]
\\ &+ \rho \,
\mathbb{E}_{\sigma_t}
\left[
\gamma(\sigma_t) \,
\mathcal{L}_{\text{repel}}\big(D_\theta(S_0 + \sigma_t \epsilon, \sigma_t, z_0)\big)
\right], \nonumber
\label{eq:full_loss}
\end{align}

where $w(\sigma_t)$ is the noise-dependent weighting term described in Appendix~\ref{app:shape_edm}.

\paragraph{Keypoint Chamfer Loss ($\mathcal{L}_{\text{chamfer}}$):}
Encourages predicted keypoints to lie near the input point cloud:

\begin{equation}
\mathcal{L}_{\text{chamfer}} = d_{\text{CD}}(K, S_0),
\end{equation}

where $K \in \mathbb{R}^{d \times 3}$ is the set of predicted keypoints and $S_0 \in \mathbb{R}^{N \times 3}$ is the input shape.

\paragraph{Deformation Consistency Loss ($\mathcal{L}_{\text{mse}}$):}
Promotes consistency of keypoints under geometric deformation:

\begin{equation}
\mathcal{L}_{\text{mse}} = \frac{1}{d} \sum_{i=1}^{d} \left\| \mathcal{T}(k_i) - k_{i}^{\text{deformed}} \right\|_2^2,
\end{equation}

\paragraph{Furthest Point Sampling Loss ($\mathcal{L}_{\text{FPS}}$):}
Used during initial training to bootstrap keypoint coverage. Given a set of predicted keypoints $K$ and a reference set $K_{\text{fps}}$ obtained by applying Furthest Point Sampling to the input point cloud:

\begin{equation}
\mathcal{L}_{\text{FPS}} = d_{\text{CD}}(K, K_{\text{fps}}),
\end{equation}

where both sets contain $d$ keypoints and $d_{\text{CD}}$ is the Chamfer Distance.

\paragraph{KL Divergence Loss ($\mathcal{L}_{\text{KL}}$):}
Regularizes the auxiliary latent code $z_{\text{aux}} \sim \mathcal{N}(\mu, \sigma^2)$ toward a unit Gaussian prior:

\begin{equation}
\mathcal{L}_{\text{KL}} = \frac{1}{2} \sum_{i=1}^{m} \left( \mu_i^2 + \sigma_i^2 - \log \sigma_i^2 - 1 \right),
\end{equation}

where $\mu \in \mathbb{R}^m$ and $\sigma^2 \in \mathbb{R}^m$ are predicted by the encoder from the input shape.

\paragraph{6. Full Training Objective:}
Combining all loss terms, the total loss is:

\begin{equation}
\mathcal{L}_{\text{total}} = \lambda_0 \mathcal{L}_{\text{FPS}} + \lambda_1 \mathcal{L}_{\text{diff}} + \lambda_2 \mathcal{L}_{\text{chamfer}} + \lambda_3 \mathcal{L}_{\text{mse}} + \lambda_4 \mathcal{L}_{\text{KL}},
\end{equation}

where the weighting coefficients $\lambda_4$ is tuned over the training schedule.

\section{Limitations and Broader Impact}
Our KeyPointDiffuser model automatically identifies semantically meaningful keypoints without requiring any human annotations, eliminating the need for manual labeling. It leverages diffusion models to reconstruct higher-quality point clouds and is evaluated solely on publicly available datasets. As a result, our method has very limited negative societal impact, since it avoids the use of sensitive or proprietary data and does not rely on human inputs that could introduce bias or privacy concerns.

\section{Ablation Study: Varying Number of Keypoints}
\label{app:num_keypoints}

We further investigated the airplane category. To understand this behavior more comprehensively, we conducted an ablation across different numbers of keypoints (ranging from 10 to 128), evaluating performance using keypoint correlation and DAS. The result is shown in Table~\ref{tab:corr_das_keypoints}.

The baselines show different stability behaviors, particularly in DAS. KPD and SC3K maintain reasonably strong correlation across keypoint counts but exhibit noticeable declines in DAS, dropping by roughly 18 and 15 percentage points respectively, indicating reduced spatial coherence at higher resolutions. KeyGrid tends to produce more stable results for the DAS metric. Skeleton Merger performs well at lower keypoint counts but does not scale computationally, because its predicted skeleton parts grow quadratically with the number of keypoints, the method becomes increasingly inefficient, and training failed to produce meaningful results even after 48 hours, leading us to terminate the high-keypoint runs.

Across all settings, our method demonstrates the strongest robustness. It consistently achieves the highest correlation and competitive DAS values, including at 128 keypoints, without exhibiting the degradation trends observed in some of the baselines. 

Interestingly, correlation tends to stay stable or even improve as the number of keypoints increases, while DAS tends to degrade with a larger number of keypoints. This is likely due to the closest-keypoint assignments in DAS being greedy and always choosing the single nearest match, while correlation assigns to all classes within a threshold distance, which is a more flexible strategy. This allows correlation to handle keypoints at boundaries more generously, since a keypoint that lies between two keypoint labels could be assigned to either one by DAS, while correlation would assign it to both part classes. This situation tends to become more common as the number of keypoints increases.

\begin{table*}[t]
\centering
\begin{adjustbox}{max width=\linewidth}
\begin{tabular}{c|ccccc|ccccc}
\toprule
\multirow{2}{*}{\# Keypoints} & \multicolumn{5}{c|}{Correlation $\uparrow$} & \multicolumn{5}{c}{Dual Alignment Score $\uparrow$} \\
& KPD & KeyGrid & SC3K & SM & Ours & KPD & KeyGrid & SC3K & SM & Ours \\
\midrule

10 & 0.9 & 0.92 & 0.83 & 0.89 & \textbf{0.96} & 0.73 & 0.76 & 0.64 & \textbf{0.8} & 0.79 \\
16 & 0.91 & 0.91 & 0.88 & 0.91 & \textbf{0.93} & 0.66 & 0.71 & 0.56 & 0.74 & \textbf{0.76} \\
32 & 0.92 & 0.88 & 0.86 & 0.91 & \textbf{0.94} & 0.67 & 0.63 & 0.55 & \textbf{0.73} & 0.70 \\
64 & 0.93 & 0.88 & 0.84 & -- & \textbf{0.96} & 0.60 & 0.62 & 0.52 & -- & \textbf{0.71} \\
128 & 0.94 & 0.95 & 0.81 & -- & \textbf{0.96} & 0.55 & \textbf{0.71} & 0.49 & -- & \textbf{0.71} \\
\bottomrule
\end{tabular}
\end{adjustbox}
\caption{Best correlation and DAS per model for varying number of keypoints}
\label{tab:corr_das_keypoints}
\end{table*}

\section{Ablation Study: Effect of Loss Terms}
\label{app:loss}
To better understand the contribution of each component in our method, we conduct a series of ablation studies by systematically removing loss, presented in Table~\ref{tab:ablation_losses}. These experiments are performed on the Airplane category and evaluated using keypoint correlation and DAS.

\begin{table}[t]
\centering
\begin{adjustbox}{max width=\linewidth}
\begin{tabular}{l|cc}
\toprule
Zeroed Loss Term & Correlation ↑ & DAS ↑ \\
\midrule
None (full model) & 0.96 & \textbf{0.79} \\
$\lambda_0 = 0$ (FPS Loss) & \textbf{0.97} & 0.75 \\
$\lambda_1 = 0$ (Diffusion Loss) & 0.96 & 0.73 \\
$\lambda_2 = 0$ (Chamfer Loss) & 0.95 & 0.77 \\
$\lambda_3 = 0$ (Deformation Consistency Loss) & 0.95 & 0.74 \\
$\lambda_4 = 0$ (KL Loss) & 0.96 & 0.77 \\
\bottomrule
\end{tabular}
\end{adjustbox}
\caption{Ablation study: effect of zeroing each loss term on correlation and DAS metrics.}
\label{tab:ablation_losses}
\end{table}

Zeroing individual loss terms highlights their distinct contributions to both correlation and DAS. Removing the Chamfer loss ($\lambda_2 = 0$) produces the strongest degradation in correlation (from 0.960 to 0.947), underscoring its importance in aligning predicted keypoints with input geometry. The diffusion loss ($\lambda_1 = 0$) also causes a noticeable drop in performance, reducing both correlation and DAS, suggesting the strong role of reconstruction to improve keypoint performance. Zeroing the deformation consistency loss ($\lambda_3 = 0$) leads to a moderate decline in both metrics, indicating that it helps maintain stable semantic structure under deformation. Interestingly, removing the FPS loss ($\lambda_0 = 0$) slightly improves correlation, though with a trade-off in DAS. The KL loss ($\lambda_4 = 0$) has a mild effect on both metrics, suggesting that while it stabilizes the latent space, the model remains relatively robust without it.

\section{Latent Space Dimensionality}
\label{app:latent}

To understand the effect of the size of $z_{aux}$ more comprehensively, we conducted an ablation across different sizes of the latent dimension (ranging from 5 to 20) on the mug category, evaluating performances in both reconstruction and evaluation quality.

\begin{table}[h]
\centering
\begin{adjustbox}{max width=\linewidth}
\begin{tabular}{c|c|c|c}
\hline
\textbf{Latent Space Size} & \textbf{Chamfer Distance} & \textbf{EMD} & \textbf{MMD-CD} \\
\hline
5  & 0.016 & 3.7e-05 & 0.010  \\
10 & 0.011 & 3.5e-05 & 0.012 \\
15 & 0.008 & 3.6e-06 & 0.015 \\
20 & 0.007 & 2.9e-06 & 0.017 \\
\hline
\end{tabular}
\end{adjustbox}
\caption{Varying the size of the latent space improves the reconstruction, however, degrades the generation results.}
\end{table}

As the auxiliary latent space dimensionality increases, the model gains more representational capacity. This leads to better reconstruction performance, which is reflected in consistently decreasing Chamfer Distance and EMD scores for larger latent sizes. A higher-dimensional auxiliary latent code allows the autoencoder to store more geometric detail from each input, thereby reducing reconstruction error. However, generation quality does not improve in the same way. During generation, the latent vector is obtained by taking the mean latent distribution from the training set, meaning it is likely to provide less information.

\section{EgoBody: Data Generation}

We convert SMPL-X parameters from EgoBody~\cite{zhang2022egobody} into a standardized geometric and semantic representation consisting of a surface point cloud, sparse structural keypoints, and four-way body-part labels. SMPL-X parameters are first passed through the SMPL-X model to obtain a mesh and joint locations. From this mesh, we uniformly sample 2048 surface points to obtain a point-cloud representation. We additionally extract a compact set of keypoints formed from major anatomical joints (pelvis, chest, head, shoulders, elbows, wrists, hips, knees, ankles) together with long-part midpoints (e.g., upper arm, forearm, thigh, shin, torso). For semantic part     labeling, we interpolate linear blend skinning (LBS) weights at each sampled point and aggregate joint influences into four regions: torso, head, arms, and legs. These assignments are then refined using two heuristical priors. (1) Limb proximity prior: for each arm and leg, we compute the minimum distance from a point to the corresponding long bone segments (e.g., shoulder–elbow and elbow–wrist for arms; hip–knee and knee–ankle for legs). Points close to these segments receive an exponential boost for the associated limb class, which suppresses torso spillover around the shoulders and hips. (2) Head–torso boundary prior: to prevent ambiguous transitions around the neck, we construct a separating plane defined by the bisector of the chest to head and chest to pelvis directions, and additionally evaluate each point’s radial distance to the head axis (chest to head). Points lying on the head-facing side of this plane and within a reasonable cylindrical radius are promoted to the head class, stabilizing the boundary even when LBS weights alone are unreliable. Final labels are obtained after K-NN mode smoothing. The pipeline outputs a point cloud accompanied by a compact keypoint representation and four-way semantic class labels.

\section{EgoBody: Experiments}

Standard DAS assigns each predicted keypoint to the single nearest annotated keypoint. Because the EgoBody annotations are relatively dense, several annotated keypoints often lie at similar distances to a predicted keypoint. This makes the standard DAS sensitive to small local perturbations and produces unstable scores. To obtain a more stable evaluation, we modify DAS by allowing a predicted keypoint to match any annotation within a relative distance window of $\pm 10\%$ around the nearest keypoint distance. Concretely, if the closest annotated keypoint lies at distance $d_{\min}$, then all annotations within $[d_{\min}, 1.1, d_{\min}]$ are considered valid matches. This softens the decision boundary and makes the metric behavior more aligned with the correlation measure, which already aggregates over local neighborhoods.

Our method achieves the highest score among all baselines (Table~\ref{tab:people}), indicating that our predicted keypoints are not only geometrically consistent but also semantically stable across subjects. The improvements in both the correlation metric and DAS demonstrate that our approach produces keypoints that align more reliably with meaningful body parts and maintain their semantic identity across different people.

\begin{table}[t]
\centering
\begin{adjustbox}{max width=\linewidth}
\begin{tabular}{c|ccccc}
\toprule
Metric & KPD & KeyGrid & SC3K & SM & Ours \\
\midrule
Correlation $\uparrow$ & 0.682 & 0.950 & 0.815 & 0.882 & \textbf{0.988} \\
DAS $\uparrow$          & 0.454 & 0.533 & 0.472 & 0.504 & \textbf{0.623} \\
\bottomrule
\end{tabular}
\end{adjustbox}
\caption{Correlation and DAS on the EgoBody dataset. Our model outperforms others approaches in both metrics    }
\label{tab:people}
\end{table}

\begin{figure}
    \centering
    \includegraphics[width=0.25\linewidth,height=0.35\linewidth]{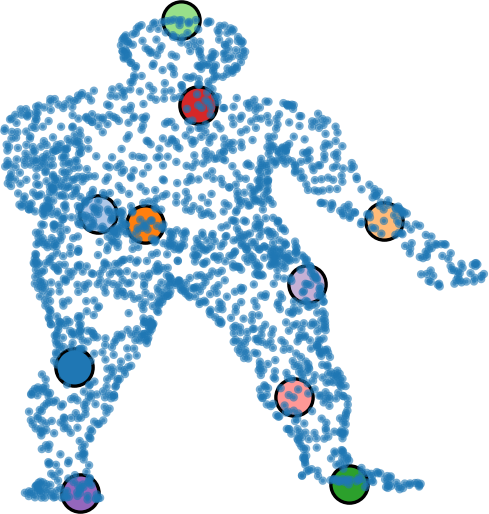}
    \includegraphics[width=0.25\linewidth,height=0.35\linewidth]{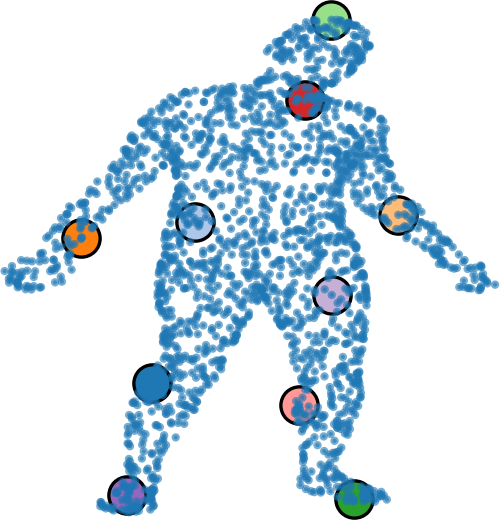}
    \includegraphics[width=0.25\linewidth,height=0.35\linewidth]{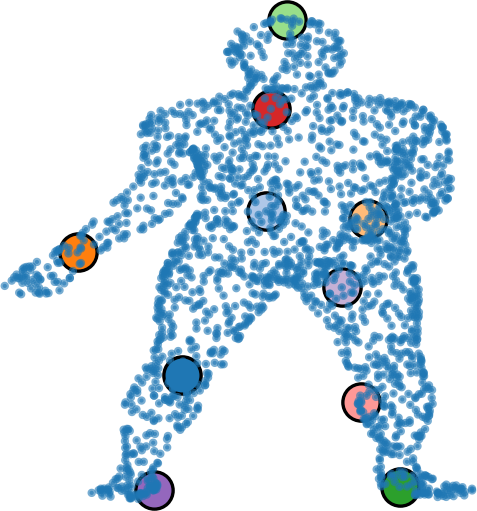}
    \caption{Qualitative results on the EgoBody dataset. We visualize predicted keypoints (colored markers) overlaid on the input point clouds. The keypoints remain stable across subjects and poses, highlighting strong semantic consistency.}
\end{figure}

\end{document}